# Autonomous Construction-Site Safety Inspection Using Mobile Robots: A Multilayer VLM-LLM Pipeline

Hossein Naderi[1], Alireza Shojaei[2], Philip Agee[3], Kereshmeh Afsari[4], Abiola Akanmu[5]

[1]PhD Candidate, Dept. of Building Construction, Myers-Lawson School of Construction, Virginia Tech, Blacksburg, VA. ORCID: https://orcid.org/0000-0002-6625-1326. Email: hnaderi@vt.edu

[2]Assistant Professor, Dept. of Building Construction, Myers-Lawson School of Construction, Virginia Tech, Blacksburg, VA (corresponding author). ORCID: https://orcid.org/0000-0003-3970-0541. Email: shojaei@vt.edu

[3]Assistant Professor, Dept. of Building Construction, Myers-Lawson School of Construction, Virginia Tech, Blacksburg, VA, email: pragee@vt.edu

[4]Assistant Professor, Dept. of Construction Engineering and Management, Myers-Lawson School of Construction, Virginia Tech, Blacksburg, VA, email: keresh@vt.edu

[5]Associate Professor, Dept. of Construction Engineering and Management, Myers-Lawson School of Construction, Virginia Tech, Blacksburg, VA, email: abiola@vt.edu

**Abstract**

Construction safety inspection remains mostly manual, and automated approaches still rely on task-specific datasets that are hard to maintain in fast-changing construction environments due to frequent retraining. Meanwhile, field inspection with robots still depends on human teleoperation and manual reporting, which are labor-intensive. This paper aims to connect what a robot sees during autonomous navigation to the safety rules that are common in construction sites, automatically generating a safety inspection report. To this end, we proposed a multi-layer framework with two main modules: robotics and AI. On the robotics side, SLAM and autonomous navigation provide repeatable coverage and targeted revisits via waypoints. On AI side, a Vision Language Model (VLM)-based layer produces scene descriptions; a retrieval component powered grounds those descriptions in OSHA and site policies; Another VLM-based layer assesses the safety situation based on rules; and finally Large Language Model (LLM) layer generates safety reports based on previous outputs. The framework is validated with a proof-of-concept implementation and evaluated in a lab environment that simulates common hazards across three scenarios. Results show high recall with competitive precision compared to state-of-the-art closed-source models. This paper contributes a transparent, generalizable pipeline that moves beyond black-box models by exposing intermediate artifacts from each layer and keeping the human in the loop. This work provides a foundation for future extensions to additional tasks and settings within and beyond construction context.

**Keywords**: Construction Safety, Construction robotics, Safety Inspection, LLMs, VLMs, AI

1. **Introduction**

With accounting for 21% of all deaths among US workers, construction industry is one of the most dangerous industries [1]. On-site safety inspections have long been a traditional method for reducing fatalities and addressing safety challenges [2]. However, these inspections rely heavily on manual efforts, which are associated with several challenges, including inaccuracy and high cost [3]. Additionally, continuous monitoring is difficult to maintain, as personnels should physically observe the site [4], requiring frequent visits to thoroughly identify potential safety hazards. Another significant challenge is the unreliability in hazard recognition, which can be inconsistent depending on the experience and expertise of the inspectors [5]. To overcome these limitations, automated safety inspection has been explored as one of the promising solutions.

With the rapid growth of sensing technologies in recent decades, various technologies are utilized in studies for automated safety inspection. Some studies use sensor-based systems for real-time alert of overhead hazards [6]. However, it requires sensors installed on construction entities and around predefined hazard, which is not suitable in a rapidly changing construction sites [7]. Additionally, the need for constant reconfiguration of sensors to adapt to shifting environments can lead to inefficiencies. Therefore, more flexible and adaptive solutions are needed to ensure safety in dynamic construction environments.

Another group of studies utilize Unmanned Aerial Vehicle (UAVs) and Quadruped Robots equipped with cameras for automated inspection. These mobile devices move faster than humans and can access hard-to-reach places and perform continuous monitoring [8], [9]. These studies can be categorized into two overall groups that each of them has its own challenges. (1) First group of study utilize the UAVs and Quadruped robots with a camera to reduce human transmission to hazardous work area and use tele-operation to recognize the hazardous area and also generate the report and do actions. For example, Halder and et al. [10] utilized Quadruped Robots and proposed a novel process for automated inspection with an inspector assistant. (2) Second group of studies equipped Robots with computer vision techniques to automate one specific task in construction inspection. For example, Gheisary et al. [11] employed computer vision techniques to detect unguard openings from video streams transmitted by UAVs. Kim et al. [12] also applied computer vision techniques for object detection and object tracking for proximity monitoring.

Despite the valuable contributions of both groups of studies, we identified knowledge gaps in both groups of studies. The first group of studies, which aims to reduce manual work through tele-operated quadruped robots and UAVs, still relies heavily on human intervention for controlling the robots, identifying inspection issues, generating reports, and taking corrective actions. The second group, which improves the recognition process using conventional computer vision techniques, also presents limitations that need further exploration. (1) Traditional computer vision methods require extensive, diverse image datasets, but construction companies are often reluctant to share site-specific images, limiting the potential of these techniques. (2) These methods are typically trained on narrowly defined, task-specific datasets, making them less effective in dynamic, rapidly changing environments. They require frequent re-training and adjustments for a broad range of safety inspection tasks. (3) Although these methods reduce dependence on manual inspection, they still require human intervention for controlling robots and UAVs, as well as for reviewing and producing inspection reports.

To address these gaps, this paper has defined three following research objectives. First, design a multi-layer framework that links scene description, regulation grounding, safety assessment, and report generation so that a mobile robot can perform automated safety inspection end-to-end. Second, implement a working prototype to validate feasibility in a realistic lab setting that mirrors construction safety settings. Third, conduct experiments to evaluate performance across multiple scenarios using standard metrics (recall, precision, accuracy, and F1), and analyze errors to guide improvement. The contributions are as follows: (1) introduction of a regulation-aware approach that detects safety violations by tying visual evidence to OSHA-based rules and producing traceable, cited outputs; (2) strong recall with competitive precision across scenarios, prioritizing missed-hazard reduction while keeping false alarms manageable; and (3) a transparent, auditable pipeline whose intermediate outputs make the reasoning easy to follow, while remaining zero-shot and modular so it can adapt to new tasks without task-specific retraining. Together, these advances move automated inspection closer to practical use in the AEC industry.

The rest of the paper is organized as follows. Section 2 reviews related studies in construction safety inspection, robotics, and vision-language models. Section 3 presents the proposed framework and its four layers. Section 4 details the implementation, including hardware, software, and rule sources. Section 5 describes the experiment design, scenarios, and metrics. Section 6 reports the results. Section 7 provides the discussion, including significance, a step-by-step analysis of a sample output, and limitations with future directions. Section 8 concludes the paper.

## 2. Related Work

### 2.1. Automated safety management in AEC

Early automation efforts in construction safety can be traced back to almost 18 years ago, when Wang and et al. [13] utilized project models to identify safety requirements for projects. These efforts were followed by other works that applied fixed or body-worn sensors [14] to track workers and equipment and trigger proximity alerts. These systems help with struck-by risk and have been validated in field and lab settings, but they need anchors, careful placement, and periodic calibration, which raises cost and creates blind spots when the site layout changes [15]. Fixed-anchor UWB can reach good ranging precision, yet performance degrades with multipath and occlusions on dynamic jobsites, and installation choices strongly affect accuracy over time [16]. BLE-based proximity alerts reduce hardware costs but are sensitive to the placement and indoor geometry, so coverage must be re-tuned as spaces and materials change [17]. Moreover, RFID-based warning and interlock demos show feasibility (e.g., excavator shutdown near a tagged worker), but they also document configuration burden and the need for continuous maintenance [18].

As these limits became clearer, some studies utilized mobile robotic platforms and UAVs for inspection. Teleoperation and "inspector-assistant" use of quadrupeds and UAVs reduce human exposure and extend reach, but they still ask a human operator, choose viewpoints, and synthesize findings. Quadruped-based remote monitoring demonstrates reliable real-time video capture and progress checks, yet path planning and inspection targeting remain operator-driven. A systematic review of robots for inspection and monitoring likewise highlights the promise of legged platforms while emphasizing the ongoing reliance on human operators for mission setup and interpretation [19]. Moreover, UAVs provide quick coverage and access to high or obstructed areas, but their integration on active sites introduces safety and operational constraints that limit continuous, unattended use [20]. To address these challenges, Simultaneous Localization and Mapping (SLAM) -based approaches were deployed in many robotic solutions, which can localize and map in unknown environments [21], [22], [23].

### 2.2. Computer Vision for Construction Safety

With the advances in Artificial Intelligence (AI), computer vision techniques were deployed within the context of construction safety because cameras see what inspectors see, and models can check those images quickly without adding extra hardware [24]. Task-specific models have tried to address Personal Protective Equipment (PPE) compliance, fall hazards, proximity risks, and housekeeping [24]. Recent studies of vision and deep learning for site safety demonstrate the trend, reporting near-real-time detection for helmets, vests, and glasses in varied lighting and viewpoints [25]. Beyond wearables, CCTV-based proximity monitoring measures worker-equipment distances and flags unsafe encroachments from a single fixed camera without extra tags [26]. Additionally, UAV-assisted proximity monitoring is applied to combine on-board detection with localization to estimate distances from above [27]. Also, studies applied photogrammetry and CV to identify fall risks and unsafe edges or openings from aerial imagery, expanding coverage while keeping people out of danger zones [28]. Lim and et al. [29] trained Convolutional Neural Network models to categorize messy zones and inform cleanup priorities and change-detection networks to push toward identifying poor housekeeping with better explanations. However, surveys repeatedly note data bottlenecks, site-to-site domain shift, and retraining overhead for task-specific models as main barriers for companies with limited labeled data or rapidly changing conditions to deploy such systems [30]. These limits motivated the use of pre-trained and zero- or few-shot models for the context of construction safety.

### 2.3. Foundation models in the AEC

Foundation models are large neural networks trained on broad, diverse data so they can be adapted to many downstream tasks with little or no task-specific training; this family includes text-only LLMs and multimodal vision-language models (VLMs) [31]. In AEC, early uses show how these models can help connect site imagery and safety knowledge to practical workflows. LLMs have been utilized to extract and organize construction safety regulations and answer safety queries for practitioners, turning standards into searchable guidance [31]. One other study also deployed LLMs to simulate a game-based training system for enhancing the learning process within the context of construction safety [32]. Prompt-based systems also applied to translate building codes into structured checks for semi-automated compliance with high reported precision and recall [33]. On the vision side, recent studies frame safety guidelines as prompts to evaluate VLMs on construction images for hazard identification, reporting promising zero-shot results while noting latency in real-time use [34]. Other work leverages image-language embeddings to turn site photos into draft safety observations, pointing to a path from raw imagery to report text [35]. These features speak directly to the problems outlined earlier. Computer-vision pipelines that need large, site-specific datasets face domain shift and frequent retraining. On the other hand, foundation models support zero-/few-shot prompting and can generalize across tasks and sites with far less labeling. Teleoperated UAVs and quadrupeds reduce exposure but leave report writing and rule lookups to people; LLMs with retrieval-augmented generation can ground findings in OSHA or company policies and draft cited explanations, easing the documentation workflow. In this paper, we apply these capabilities by using a VLM for scene descriptions, an LLM with retrieval for regulation-aware reasoning, and an explicit safety assessment layer that turns evidence into reports.

Bringing these strands together, prior approaches fall into two groups. (1) Teleoperated robots and manual video review reduce exposure but leave humans to plan paths and write reports, which caps scalability as site size grows. (2) Narrow, task-specific detectors achieve strong single-task metrics but depend on curated datasets and struggle to generalize across sites and seasons without retraining. Also current Foundation models approaches have three main limitations: (1) First, single-image VLMs still struggle with step-by-step reasoning about object and tying them into the validated context (like safety rules) [36]; (2) many frontier-scale foundation models are proprietary and typically accessed through cloud APIs, which is a poor fit for data-sensitive AEC deployments and for sites with limited or unstable connectivity. Policy analyses note the concentration of capabilities in a few vendors and the operational implications of API-based access, while construction IT studies document "connected jobsite" constraints and networking barriers on active projects [37] (3) foundation models are largely black boxes, offering limited traceability for why a specific safety judgment was made; the interpretability literature argues for transparent, reviewable mechanisms in high-stakes settings and reporting tools to support accountability [38]. With these challenges in mind, this paper aims to utilize autonomous robotics and foundation models to generate safety reports.

## 3. Framework
### 3.1. Overview

This section presents the overall architecture of the proposed system, as illustrated in Figure 1, which is organized into two main modules: Module A, including the robotic components, and Module B, comprising the AI components. The framework uses three distinct visual elements to represent different types of components within the system. Oval shapes represent the hardware and sensors responsible for environmental perception and data collection, such as the RGB and LiDAR/Depth cameras. Rounded rectangles, labeled from A.1 to B.4, represent the computational units that process the data through a sequence of reasoning steps, each with a specific function discussed in the following subsections. A standard rectangle is used to represent the communication channel, specifically, WebRTC, which facilitates the transfer of data from Module A to Module B. The data flow begins with the robot collecting point cloud information using its onboard LiDAR and RGB-depth cameras, followed by SLAM processing and autonomous navigation, which together ensure spatial awareness and movement within the environment. Subsequently, the RGB camera captures image frames that are transmitted to the AI module, where VLMs generate scene descriptions (B1), LLMs derive relevant safety regulations (B2), and a second round of visual-language reasoning determines the compliance of each scene with the generated rules (B3). Finally, an LLM model consolidates all frame-level safety decisions to generate a comprehensive safety report (B4). Figure 1 thus provides a high-level view of the system architecture and data flow, offering a general understanding of the interactions between hardware and software components, while the subsequent sections provide a detailed explanation of each individual module.

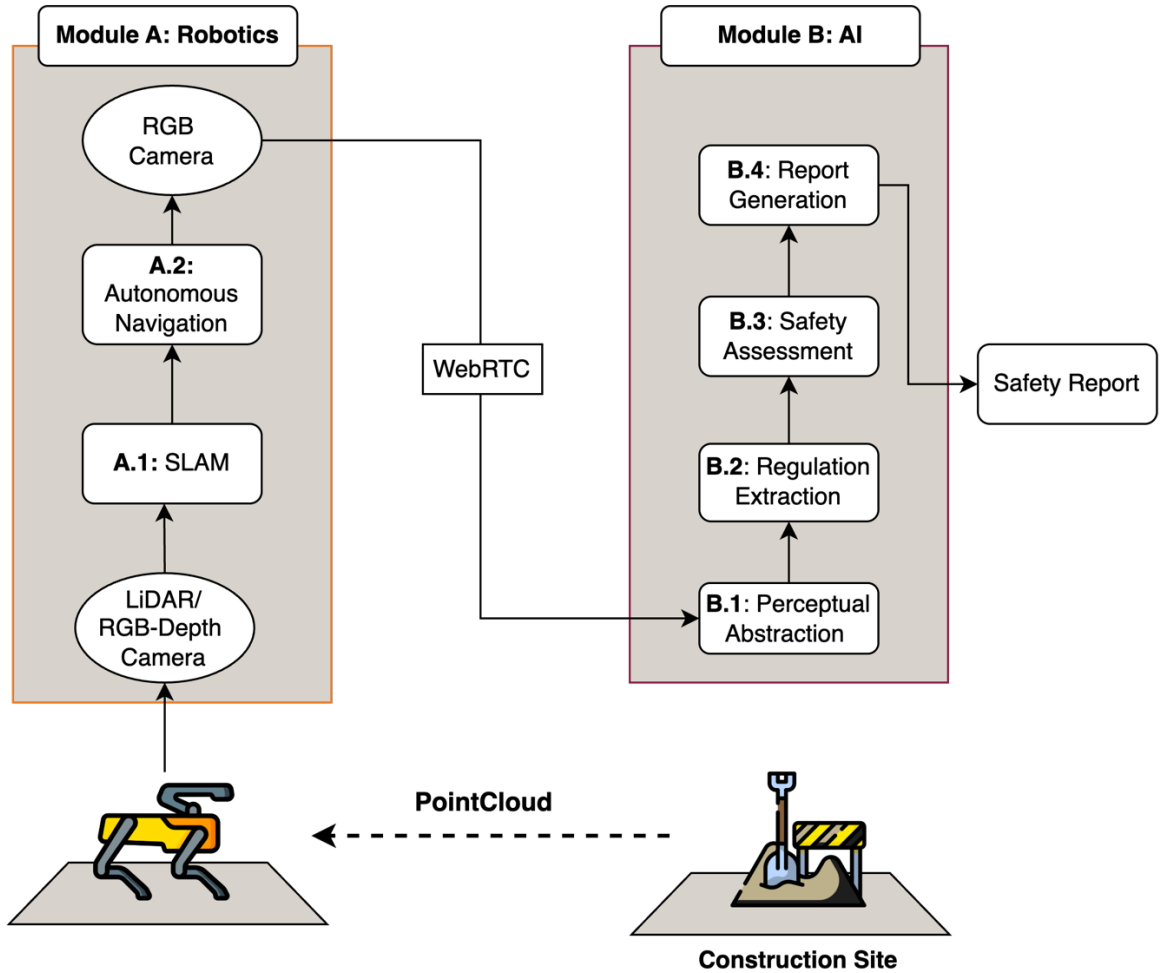

Figure 1 Overview of the framework structure and components

### 3.2. Module A: SLAM and Autonomous Navigation

Simultaneous Localization and Mapping (SLAM) and Autonomous Navigation are foundational capabilities in mobile robotics that enable a robot to move and operate independently in an environment without the need for human teleoperation [39]. From a robotics perspective, SLAM refers to the process by which a robot constructs a map of an unknown environment while simultaneously estimating its own position within that map [40], [41]. Autonomous Navigation, on the other hand, builds upon SLAM to allow the robot to plan and execute paths from one location to another safely and efficiently [42]. In the context of construction job sites, where manual robot operation is not scalable or safe, these capabilities play a vital role in enabling physical autonomy, allowing the robot to explore, inspect, or monitor areas without constant human intervention. Within our proposed framework, SLAM and navigation builds the underlying automation layer that provides spatial awareness and mobility, thereby enabling higher-level AI modules to focus on perception, reasoning, and decision-making.

Specifically, in our system, SLAM receives raw point cloud data from LiDAR and RGB-D cameras mounted on the robot. Utilizing graph-based SLAM [41], the system aligns successive sensor readings to estimate the robot's position over time and to incrementally build a spatial map of the environment. This generated map, along with the robot's real-time odometry, is passed to the Autonomous Navigation module, which utilizes behavior trees with major components of global planner, local planner, controller, and feedback monitor [43]. The global planner in our framework utilizes Dijkstra's algorithm [42] to generate a path across a global cost map. While our implementation employs Dijkstra, the framework is also compatible with other search-based planners such as A*. The local planner refines this using a local cost map to account for immediate obstacles. The controller translates these plans into specific velocity commands for the robot, and the feedback module ensures resilience by adapting to failures, for instance, re-planning paths when unexpected dynamic obstacles appear. It is important to note that while our framework utilizes standard SLAM and navigation tools, this paper does not seek to advance the state-of-the-art in those fields. Instead, they are implemented here to enable the autonomous data collection necessary for our AI-driven safety reasoning pipeline.

### 3.3. B1: Perceptual Abstraction

This component, located within the AI module, is responsible for transforming raw visual input from the robot into structured, semantically rich descriptions that can be reasoned about by subsequent components. The process begins with RGB frames streamed from the robot's onboard camera, each tagged with a precise timestamp to maintain temporal alignment with other data streams. These frames are then passed through a VLM capable of image captioning and grounded visual understanding, leveraging large-scale multimodal training to identify and describe relevant objects, activities, and spatial relationships within a construction site. First, this step is critical because it abstracts away low-level pixel and geometry data into textual descriptions, enabling downstream language models to apply reasoning and regulatory knowledge without requiring direct image interpretation. Second, the choice of VLMs allow the framework to exploit their core strengths to provide a generalizable description for dynamic environment of construction sites. As illustrated in Figure 2 the perceptual abstraction stage acts as the essential bridge between raw sensor data and the high-level reasoning pipeline that follows.

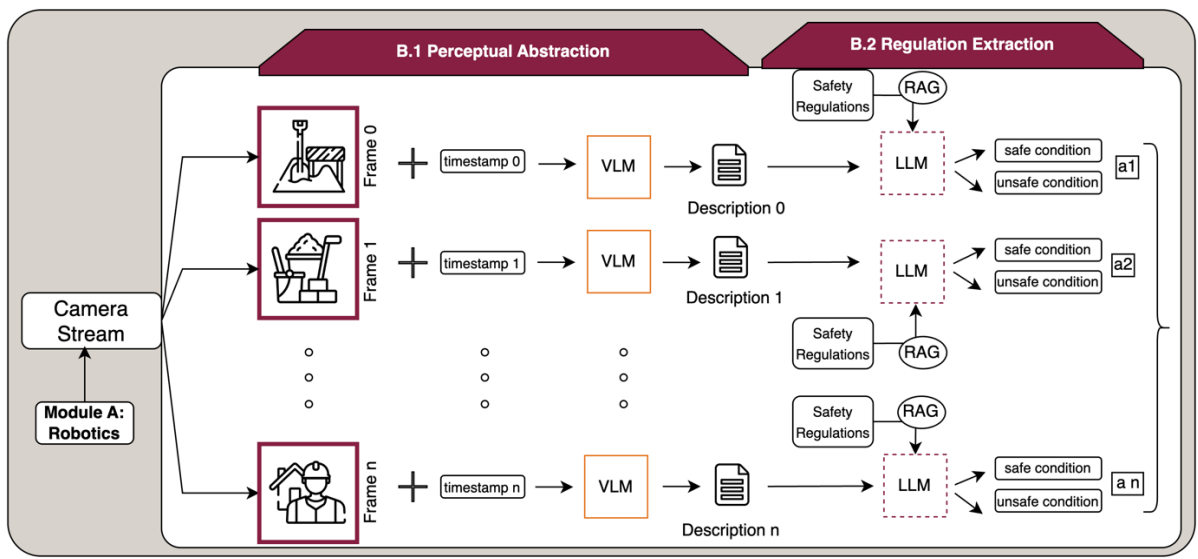

*Figure 2* Structure of Perceptual Abstraction and Regulation Generation Components in the AI Module

### 3.4. B2: Regulation Extraction

As shown in Figure 2, the Regulation Extraction component plays an important role in incorporating context-awareness into the subsequent safety assessments sections by grounding the reasoning process in established safety regulations. Without incorporating such domain knowledge, downstream models would rely solely on visual descriptions and their training datasets, which mostly over rely on parametric, and static knowledge and lacks updated regulations [44]. This challenge potentially overlooks safety-critical conditions that are not visually obvious but are well-defined in safety standards. In this stage, the textual scene descriptions produced by the Perceptual Abstraction module (section 3.3) are combined with a set of construction safety regulations. These inputs are processed using a Retrieval-Augmented Generation (RAG) component, which first retrieves the most relevant regulatory clauses (or chunks) from a safety knowledge base (or embedding database) and then feeds them into a LLM for reasoning [45]. RAG is employed for two main reasons: (1) it ensures that the LLM operates with up-to-date and domain-specific safety knowledge without the need for costly retraining, and (2) it constrains the LLM's reasoning to verifiable sources, reducing the risk of hallucination. The LLM is responsible for interpreting the retrieved regulations in the context of each frame's description, producing a structured output that labels potential environment (construction site) as (a) safe or (b) unsafe. Apart from the necessity of mentioning unsafe conditions, it's important to provide the context regarding the safe condition as the language model generally relies on the context and may misinterpret without it [46]. This regulatory grounding not only improves the reliability of the safety assessment process (backed with RAG component) but also ensures that every classification is traceable back to a safety rule.

### 3.5. B3: Safety Assessment

As illustrated in Figure 3, the Safety Assessment component leverages the contextual information provided by the previous Regulation Generation module to determine whether each visual scene captured by the robot meets or violates safety requirements. This stage is central to the framework, with all prior modules serving as foundational steps to ensure that the assessment is both accurate and contextually grounded. Each frame is paired with the corresponding set of safety and unsafe conditions generated earlier, creating a rule-aware context for analysis. The combined input is then processed by a VLM, which evaluates the frame against these conditions to produce a safety classification, identifying whether the depicted scene is compliant or non-compliant. In addition, each assessment is linked to a precise timestamp, ensuring temporal traceability and enabling the integration of sequential safety states in the final reporting stage. This timestamping is essential for the last module, where the system summarizes the frame-by-frame assessments into a coherent safety report.

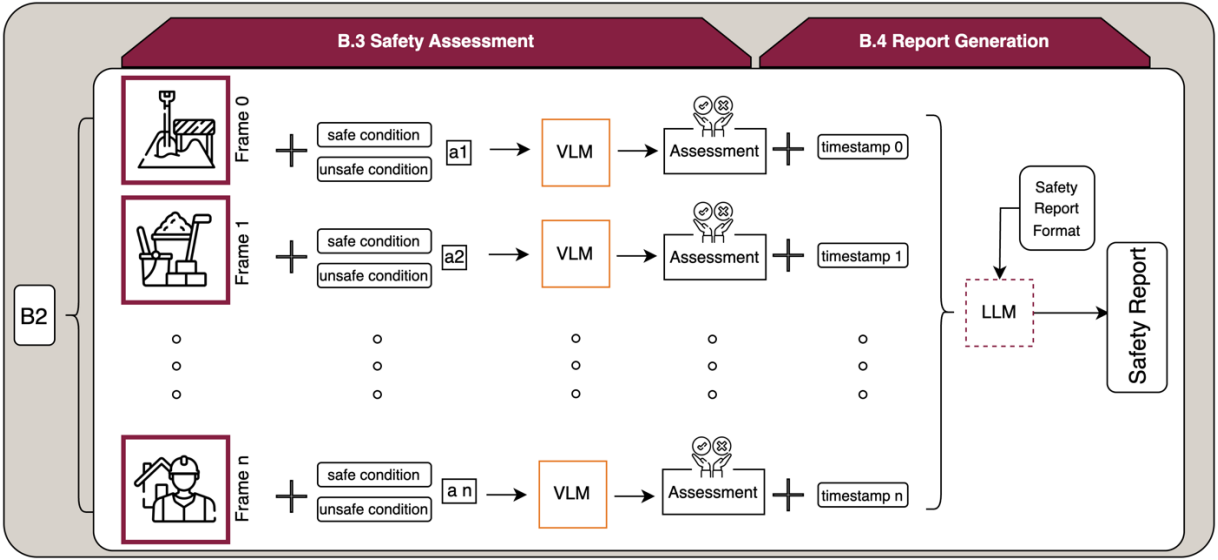

*Figure 3* Structure of Perceptual Abstraction and Regulation Generation Components in the AI Module

### 3.6. B4: Report Generation

This module is separated from the Safety Assessment stage, as prior research has shown that LLMs tend to perform best when focused on a single reasoning task at a time rather than handling multi-step visual reasoning and long-form summarization simultaneously [47]. By isolating this stage, the framework allows earlier modules to specialize in perception, rule grounding, and safety classification, while reserving the LLM in this module for its core strength, synthesizing coherent, contextually rich narratives from structured inputs [48]. In this step, the LLM receives the sequence of timestamped safety assessments for each frame, enabling it to produce a safety report that not only documents the safety status of each observed moment but also captures patterns and temporal relationships across the entire observation period. This temporal context serves two purposes: first, it supports more accurate and insightful reporting by identifying recurring hazards, and second, it enables the LLM to adapt the content into multiple output formats as needed (e.g., structured logs, formal inspection reports, or compliance checklists). The inclusion of predefined report formatting is equally critical, as it provides the LLM with structural guidance, ensuring the generated report follows the expected organization, style, and completeness standards.

## 4. Implementation

This section presents the implementation details of the proposed framework, serving as a proof of concept to validate its feasibility and prepare it for evaluation in following sections. The section is organized into three parts: first, the hardware setup and configuration; second, the implementation of the Robotics Module described in Section 3.2; and third, the implementation of the AI Module including the perception, regulation generation, safety assessment, and report generation stages. All codebases, configuration files, and related resources are publicly released under an open-source license to encourage transparency, reproducibility, and further development by the research community [1].

### 4.1. Hardware Configuration

To implement the framework, this paper employed a quadruped robot with various attachments. Figure 4 illustrates the robot and various equipment attached to it. This quadruped robot is a Unitree Go2 Edu that has a built-in RGB camera, and a K1 Robotic Arm and a Robosense Helios32 Lidar is attached to the robot. In this implementation, all components in the robotic section of framework (module A) are conducted on a NVIDIA Jetson Nano attached to the robot. Moreover, the implementation regarding the AI module (module B) was mainly performed on an Ubuntu 20.04 Linux operating system running on a high-performance workstation equipped with an AMD Ryzen Threadripper 3960X 24-core processor, 258 GB of RAM, dual Samsung 980 PRO 2 TB NVMe SSDs, and two NVIDIA GeForce RTX 3090 GPUs (each with 24 GB VRAM, driver version 550.120, CUDA 12.4).

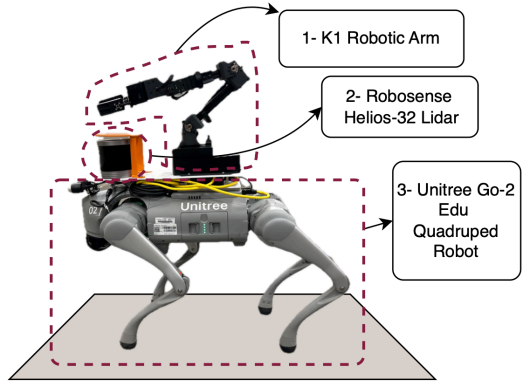

*Figure 4* Robot and its attachments

### 4.2. Robotics Implementation

Figure 5 illustrates the complete flow of the Robotics Module implementation, showing how SLAM and autonomous navigation were integrated on the Unitree Go2 Edu platform and our proposed framework. The process begins with point cloud data from the RoboSense RS-Helios-32 LiDAR and RGB images from the camera (Step 1), streamed to the NVIDIA Jetson Nano onboard the Unitree Go2 robot. This data is processed by the SLAM system (Step 2) using the RTAB-Map package [49] in ROS2 Foxy. The LiDAR driver was set up via the official RoboSense SDK, with frame IDs and ROS topics configured to ensure consistency in TF trees and message flow. The RTAB-Map configuration allowed switching between mapping mode, where the system incrementally builds an environmental map, and localization mode, where the robot localizes itself against a prebuilt map without updates. Core nodes included icp_odometry for estimating location and motion via Iterative Closest Point [50] alignment of point clouds, rtabmap for map construction and loop closure detection, and rtabmap_viz for visualization of robot trajectory and generated maps. Sensor fusion incorporated RGB and depth images, along with IMU data, to enhance loop closure detection and odometry accuracy. This SLAM configuration continuously published both /odom and /map topics, providing the foundational spatial data for the navigation system.

---

[1] https://github.com/h-naderi/unitree-go2-slam-nav2

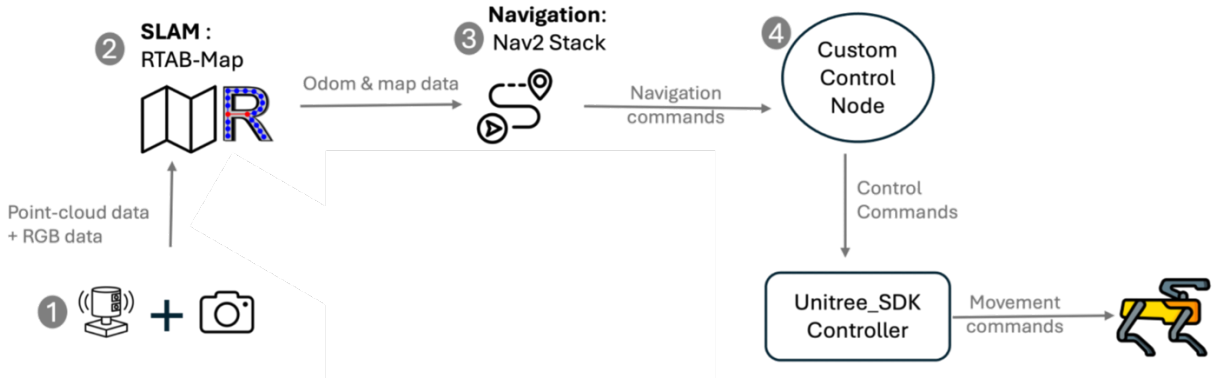

*Figure 5 The implementation process of Robotic module*

SLAM outputs odometry and map data (Step 3), which serve as inputs to the Nav2 navigation stack [51], [52]. Nav2's lifecycle-managed architecture initialized global and local planners, obstacle avoidance servers, and behavior trees for navigation. The global planner generated paths on the global costmap using Dijkstra [42], while the local planner adjusted these paths in real time using the local costmap to avoid obstacles. Navigation outputs were velocity commands on the /cmd_vel topic. These navigation commands (Step 4) are sent to a dedicated custom control node (Step 4), which is responsible for translating Nav2's generic velocity commands into instructions compatible with the Unitree control nodes. The custom control node plays a crucial bridging role between high-level navigation logic and the low-level control required to physically maneuver the robot. It subscribes to Nav2's /cmd_vel topic to receive real-time velocity commands and converts them into the correct motion instruction format (SportMode request) for the Unitree SDK. This node also implements safety and reliability features, including command timeouts to stop the robot if new commands are not received. This architecture allows the robot to navigate autonomously in mapped or partially mapped environments, dynamically avoid obstacles while the navigation goals are set by users. The SLAM implementation is illustrated in Figure 6, showing a 3D map generated from LiDAR in Window A, while the robot is moving in the research lab environment (from step 1 to step 4). In step 1, when the robot was not moved, a map is created based on limited number of point cloud that can be seen from the robot standpoint. When the robot moves, the map is incrementally built and also is aligned using RTAB-Map nodes. Moreover, Figure 7 illustrates the autonomous navigation process using occupancy maps created by Nav2 stack [51], [52]. Frame 1 in this figure shows the robot when the user sets the navigation goal, while the frame 2 and 3 presents how the robot creates a path plan and follows that to avoid the obstacle in the environment.

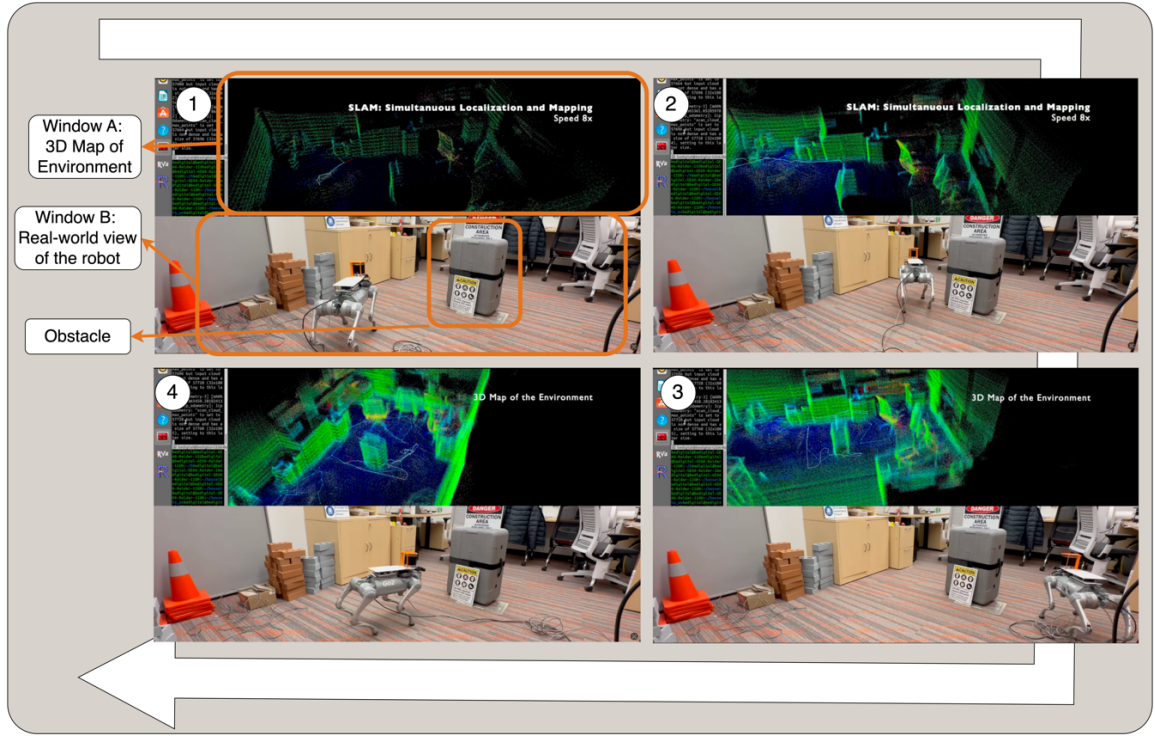

*Figure 6* SLAM Implementation

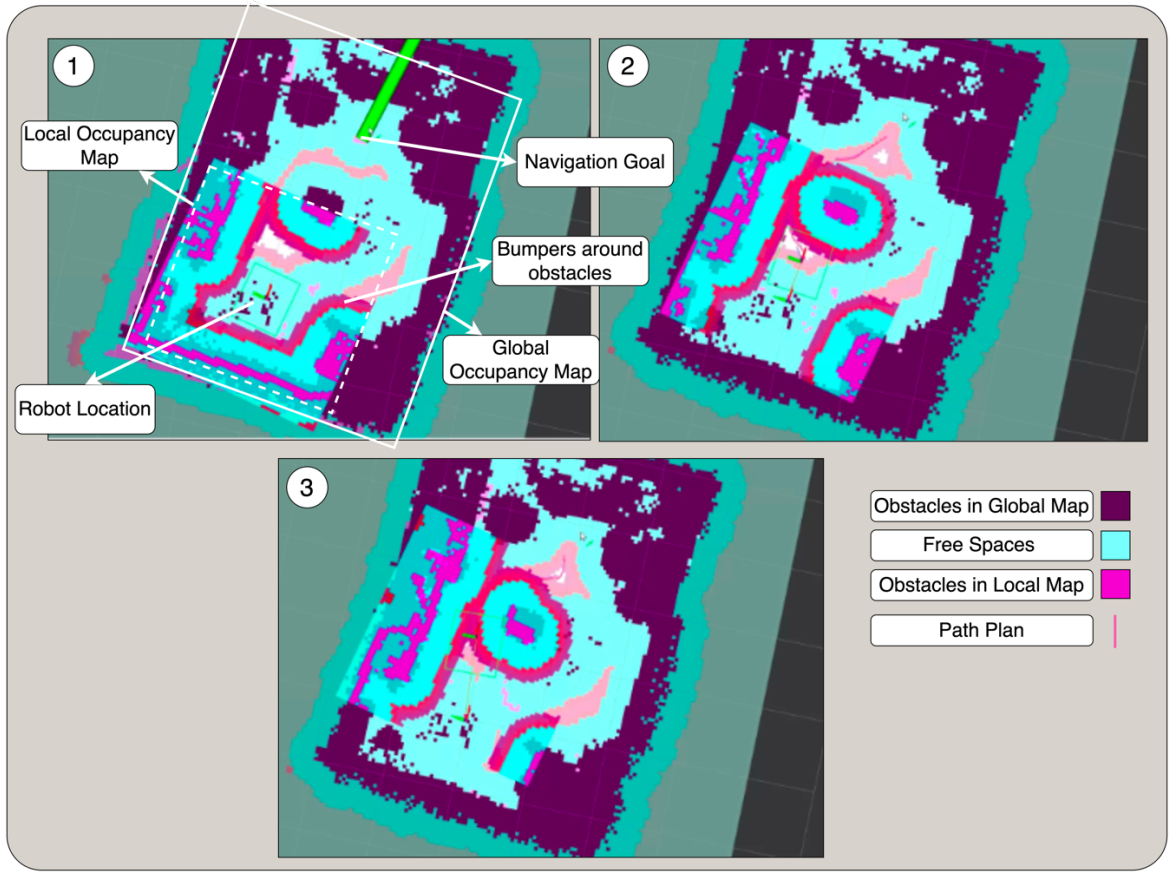

*Figure 7* Autonomous Navigation Implementation

### 4.3. Implementing the AI Module

The AI module was implemented using open-source VLMs and LLMs, selected to ensure both privacy and cost-effectiveness, which is important within the context of construction sites. In the AEC industry, companies are often hesitant to share site data such as camera feeds or incident reports, which makes cloud-based models impractical. Moreover, running open models locally eliminates this barrier while also lowering operational costs, making it more feasible for companies to integrate the system into their existing workflows. Ollama [53] was adopted as the framework because it simplifies the deployment and management of a wide range of state-of-the-art models, provides a consistent API for both VLM and LLM tasks, and supports GPU acceleration across our workstation and edge devices. This approach allowed us to iterate quickly during a one-month pre-evaluation period, testing multiple open-source VLMs (including LLaVA [54], MiniCPM-V [55], and Gemma-3 VLM [56]) and LLMs (including Mistral [57], Llama-2 [58], Gemma-2 [59], and Phi-3 [60]) under different conditions. From pre-testing stage, we selected Gemma-3 (12B) VLM for both the module B.1-perceptual abstraction and B.3-safety assessment, Llama-3.3 [61] was utilized for generating safety rules (module B.2), and a reasoning-oriented model (DeepSeek R-1 [62]) for compiling the final safety report. The reasoning model was chosen for the final stage because it consistently produced well-structured summaries and recommendations, aligning with the formatting and clarity requirements of safety documentation.

The pipeline begins by extracting frames from the input streams at a fixed sampling rate using OpenCV, with filenames embedding the exact timestamp of capture to preserve temporal context. Each frame is then processed by the Gemma-3 VLM, which is prompted to produce detailed, object-centric scene descriptions. These descriptions are passed to Llama-3.3 to generate two safe and two unsafe rules, each expressed first as an abstract safety principle and then as a measurable, concrete criterion. This dual-layer formulation is designed so that the VLM in the next stage can evaluate compliance with minimal ambiguity. In the safety assessment stage, the original frame is reintroduced alongside the generated rules to the Gemma-3 VLM, which applies a five-level severity scale to determine whether the scene is safe or unsafe. For the final reporting stage, all frame-level assessments are incorporated with their timestamps and sent to the DeepSeek-R1 reasoning model. This stage uses a fixed, template-driven prompt to produce a structured report. Figure 8 presents how different VLMs and LLMs are implemented.

*Figure 8 Sample functions for implementing AI module*

Throughout development, prompt engineering techniques [63] are utilized actively to improve the system faithfulness. To this end, four following practices are applied: (1) defining the output formats in the prompt text so the output strictly follow it (see P3 and P5 in Figure 9); (2) instruct the model to start from abstract principles and then end with measurable checks [64] to reduce vagueness and improve reasoning capabilities (see P2 in Figure 8); (3) setting standards and severity thresholds to improve objective assessment and decision making (see P4 in Figure 9); and

(4) including role instructions such as "You are a construction safety engineer" to drive outputs toward regulatory alignment (see P1 in Figure 9). For clarity in the paper, three code excerpts from our implementation will be shown: the frame extraction function with timestamped filenames, the first-pass VLM call for scene description, and the final-stage LLM prompt for safety report generation. These examples capture the essential logic of the system while keeping the presentation concise and reproducible.

```
100     llm_prompt = f"""
101     You are an expert safety engineer in construction sites.         [P1]
102     Below is a description of a scene from a construction site.
103     Based on standard safety regulations, derive two rules for a
104     safe situation and two rules for an unsafe situation.
105     For each rule, start from an abstract principle,                  [P2]
106     then explain or generalize to concrete objects/activities.
107     Your rules must be objective, valid and based on regulations.
108     It should be easy to measure with clear quantified metrics
109     for effective decision making.
110
111     Your response should be formatted:                                [P3]
112
113     "Rules for Safe Construction Situation:"
114     1. ...
115     2. ...
116
117
118     "Rules for Unsafe Construction Situation:"
119     1. ...
120     2. ...
121
122
123     Scene description:
124     {vlm_response}
125
126     Safety Regulations:
127     {context}
128     """
```

*Regulation Generation Prompt*

```
146     prompt = f"""
147     1. Examine the image and identify any potential safety hazards based on the defined rules.
148     2. Rate the severity of the hazard on a scale of 1 to 5:
149        - 1 = No real hazard or extremely minor risk                                           [P4]
150        - 2 = Mild risk that does not violate serious safety regulations
151        - 3 = Moderate risk that may need attention but is not a major legal breach
152        - 4 = Serious hazard posing high risk of harm or a clear legal breach
153        - 5 = Extremely serious, life-threatening hazard or glaring violation of the law/regulations
154
155     3. If the rating is 4 and 5, label it **Unsafe**; otherwise, if label is 1-4 it's **Safe**.
156
157
158     **Output Format**                                                                          [P5]
159     Your response must be in this format exactly:
160     Situation: [Safe or Unsafe]
161     Reason for decision: [One sentence]
162
163     **Rules:**
164     {rules_text}
165     """
```

*Safety Assessment Prompt*

*Figure 9* Prompt techniques for implementing AI module

## 5. Experiment Design

This section aims to address the primary research objective by validating the proposed framework through controlled testing. We evaluated the framework and its implementation in three designed scenarios within a real-world simulated construction lab environment based on following reasons: (1) Although there are different construction safety datasets, such as SODA dataset [65] and PPE based datasets [66], there is a lack of publicly available datasets that offer diverse safety violations and scene variations, especially when it comes to capturing from the perspective of a mobile robot in construction settings; (2) performing tests in the controlled lab environment with tailored scenarios allows for precise control over hazard placement, environmental factors, and test conditions, ensuring that all components of the framework can be systematically evaluated under known ground-truth conditions [67], [68]; (3) testing safety problems in the construction real-world environment is often logistically complex, and unsafe [69]; (4) while simulation environment, utilized in other researches [67], [68], are valuable, are less capable to fully replicate the complexities and uncertainties of real-world robotics systems. This challenge, known as the "reality gap", means that behaviors and system performance observed in simulation may not transfer accurately to physical implementations [70], [71]. Real-world testing in controlled lab settings helps bridge this gap by providing realistic, trustworthy validation.

Three scenarios were selected for validating and evaluation of the proposed framework and its implementation. Four safety violation categories were selected to represent a range of hazards that are prevalent in construction based on OSHA's regulatory framework. (1) falls [72]; (2) housekeeping/trip hazards [73]; (3) electrical safety [74]; and 4) PPE [75] violations were included because they collectively cover multiple OSHA subparts and hazard types, including the OSHA "Focus Four" categories [76], [77] of falls, struck-by incidents, caught-between, and electrocutions. Within fall hazard category, working with ladders is particularly important to consider, OSHA reports that falls from ladders account for roughly 20% of fatal falls in construction [78], and the U.S. Bureau of Labor Statistics recorded over 22,000 ladder-related nonfatal injuries in construction in 2020 alone [79]. Housekeeping and trip hazards were included because poor site housekeeping is an important factor in both falls and struck-by incidents [80]. Moreover, PPE violations were incorporated to test the framework's ability to detect conditions already discussed by traditional computer vision approaches. Table 1 lists three main scenarios and 20 specific safety violations to test the framework. Each new scenario is designed to add some complexity to test accuracy and productivity of the proposed frameworks.

*Table 1* Scenarios and associated safety violations designed

| ID | Category | Designed Safety Violation | OSHA category | OSHA reference |
|---|---|---|---|---|
| A1 | Falls / Struck-By / Caught-Between | Ladder placed where it can be displaced by adjacent equipment; not secured | Ladders (Subpart X) | 1926.1053 |
| A2 | Trip hazards | Bricks/debris scattered in walkways | Housekeeping (Subpart C) | 1926.25(a) |
| A3 | PPE | No high visible vest and no head protection while exposed to fall hazards | PPE (Subpart E) | 1926.100(a) |
| A4 | Electrocutions | Extension cord in wet area without GFCI protection | Electrical (Subpart K) | 1926.404(b) |
| A5 | Electrocutions | Flexible cord routed through water/subject to damage | Electrical (Subpart K) | 1926.405(a)(2); 1926.405(j)(1) |
| B1 | Falls / Struck-By / Caught-Between | Portable ladder leaned on a machine (unstable/unsuitable support) | Ladders (Subpart X) | 1926.1053 |
| B2 | PPE | No head protection while on ladder/under overhead hazards | PPE (Subpart E) | 1926.100(a) |
| B3 | Trip hazard | Bricks cluttered in a spot creating trip hazard | Housekeeping (Subpart C) | 1926.25(a) |
| B4 | Struck-by/ Materials handling | Bricks stored on forklift forks with load elevated/unattended | Materials handling (Subpart H) | 1926.600(a)(3); 1926.602(c)(1) |
| B5 | Trip hazard/ Material handling | Bricks stacked in the middle of a walkway (aisles not kept clear) | Materials handling (Subpart H) | 1926.602(c)(1) |
| B6 | Electrocutions | Extension cord in wet area without GFCI protection | Electrical (Subpart K) | 1926.404(b) |
| B7 | Electrocutions | Flexible cord routed through water/subject to damage | Electrical (Subpart K) | 1926.405(a)(2); 1926.405(j)(1) |
| C1 | Falls / Struck-By / Caught-Between | Working atop large machine with unprotected sides/edges (≥6 ft) | Fall protection (Subpart M) | 1926.501(b)(1) |
| C2 | Falls/ Struck-By | Using a material lift table to elevate a person (not designed/approved for personnel) | Hoists/material lifts (Subpart N) | 1926.552(b)(1) |
| C3 | Falls/ Struck-By | Elevated work accessed without proper ladder/stair/approved platform | Access (Subpart X) | 1926.1051(a) |
| C4 | Falls/ Struck-By/ Trip hazards | Bricks placed on lift platform near feet (sliding/falling object/trip hazard) | Materials handling & storage (Subpart H) | 1926.250(a)(1) |
| C5 | Struck-by/ Materials handling | Bricks stored on forklift forks with load elevated/unattended | Materials handling (Subpart H) | 1926.600(a)(3); 1926.602(c)(1) |
| C6 | Trip hazard/ Material handling | Bricks stacked in the middle of a walkway (aisles not kept clear) | Materials handling (Subpart H) | 1926.602(c)(1) |
| C7 | Electrocutions | Extension cord in wet area without GFCI protection | Electrical (Subpart K) | 1926.404(b) |
| C8 | Electrocutions | Flexible cord routed through water/subject to damage | Electrical (Subpart K) | 1926.405(a)(2); 1926.405(j)(1) |

Each scenario was designed to cover multiple violations in close-to-realistic arrangements that reflect the complexity of actual job sites. Scenarios are designed toward adding more complexity in each new stage. While the number of safety violations were increased in each new scenario (scenario A=5, scenario B= 6, scenario C= 7), the number of helping components (safety signs) are reduced. Scenario A includes three safety signs: (1) Danger- Fall Hazard, (2) Caution- Hi-vis safety vest required beyond this point, and (3) Caution- All PPE required beyond this point. Scenarios B and C only have the second sign (Hi-vis safety vest required). Exclusion of signs in scenario B and C added further complexity into safety inspection process as the exist of written violation sign in a frame at the same time with safety violation helps VLMs to better recognize the safety issue. This feature came from the nature of VLM training, where the text and images are paired to train the models [81]. Moreover, the nature of each violation in scenario B and C is changed toward less internet-spread violations to add more complexity. For example, while scenario B replicated several violations in scenario A, a less well-known configuration is used, including unsafe portable ladder leaned on a machine, materials stacked on forklift loads, and within PPE category, hi-vis vest were worn but hard hats not. Scenario C shifted this complexity, by using elevated work on unprotected large machinery, improper use of material lifts for personnel, and access without proper ladders or platforms (less well-known violations), while carrying over the clutter, material placement, and electrical hazards from previous scenarios, and exclusion of PPE violation as a well-known and easy to recognize violation. These combinations allowed the framework to be tested against both isolated hazards and overlapping conditions, simulating the mixed-risk environments typical of active construction sites.

Figures 10, 11, and 12 illustrate the experimental setups for Scenarios A, B, and C, showing the planned layouts (left) and their real-world implementations (right). In each scenario, the Unitree Go2 robot began from a designated starting point marked by label number 1 on the maroon dashed and then followed a predefined navigation path that systematically exposed it to hazards of multiple categories. As the robot navigated through the environment, it encountered hazards from multiple categories, all of which were labeled in the plan views for consistency between the designs and their implementation. Moreover, different signs provided the robot with some level of guidance in the process of inspection.

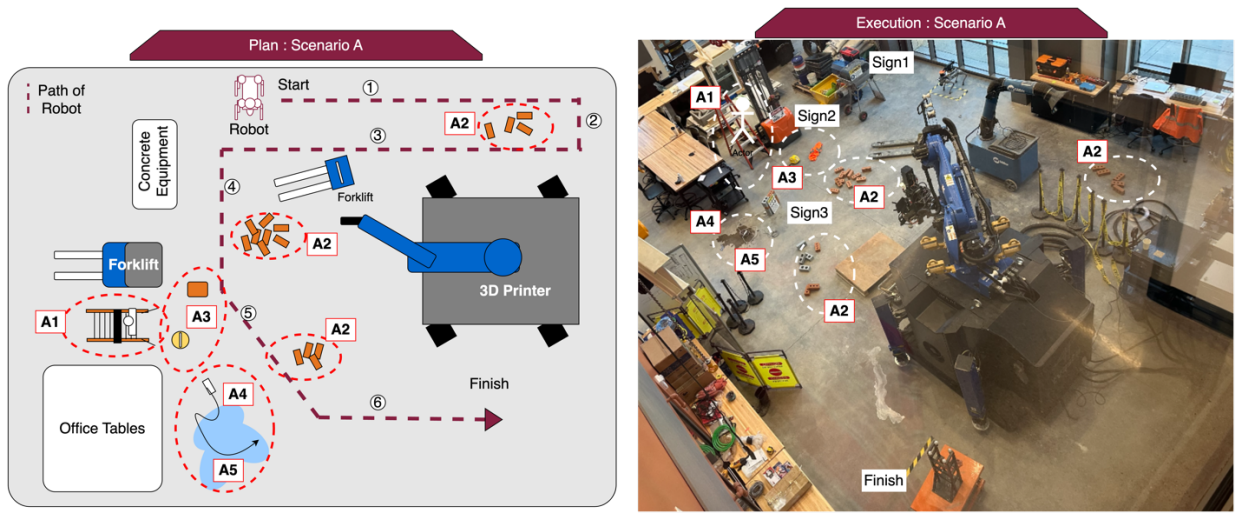

*Figure 10* Experimental setups for Scenario A

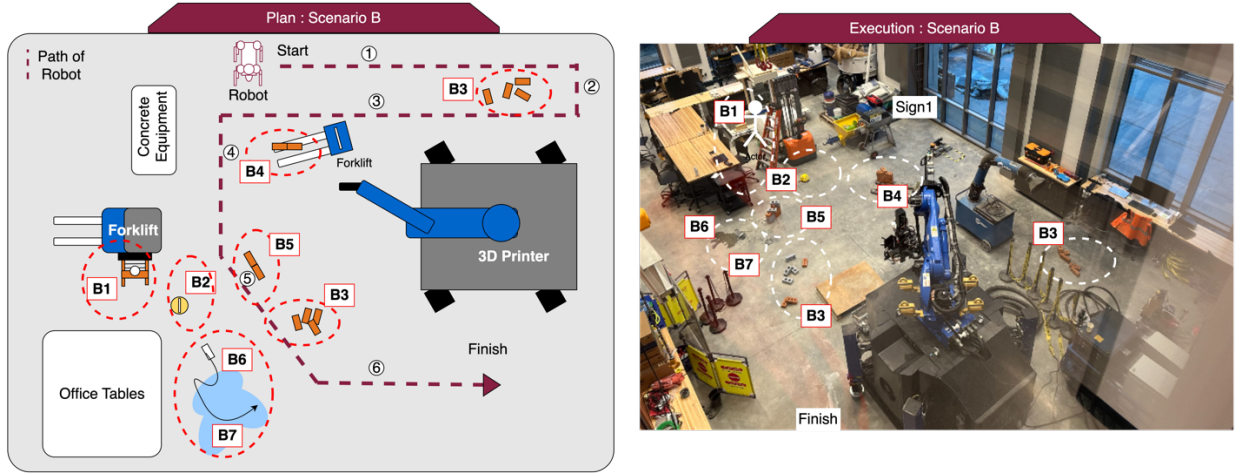

*Figure 11* Experimental setups for Scenario B

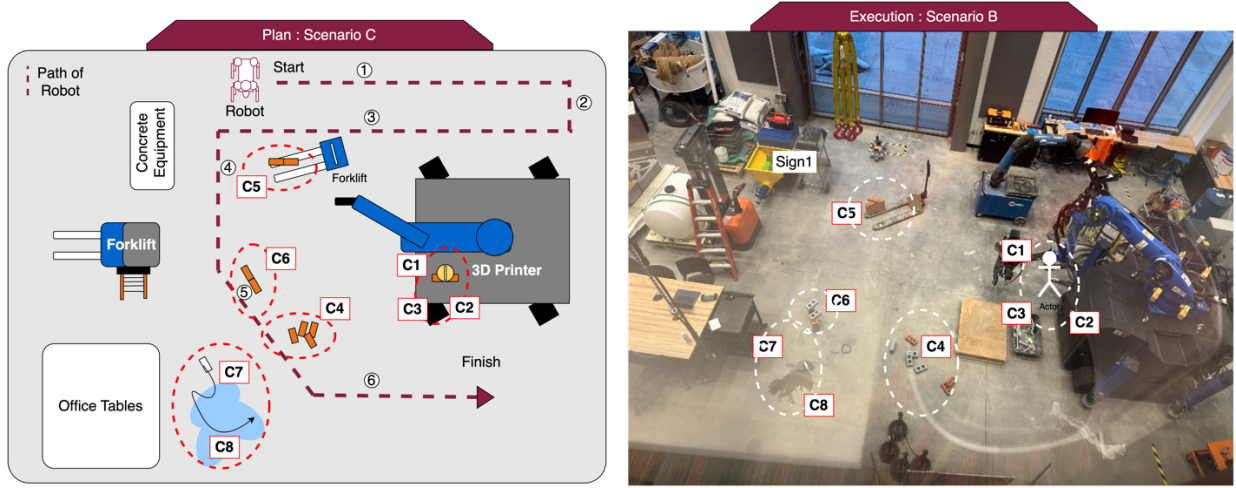

*Figure 12* Experimental setups for Scenario C

## 6. Results

The purpose of this section is to present the outcomes of our experimental study in order to evaluate the feasibility and effectiveness of the proposed framework in detecting and classifying construction safety violations. Beyond demonstrating feasibility, the goal is also to provide a reproducible benchmark for future research, allowing other scholars to compare alternative models against the same scenarios and methodology. To achieve this, we report not only the outputs of our system but also comparative results against several baseline models, highlighting both qualitative and quantitative differences. For reproducibility, the scope of our experiments and the data collected are described in detail. Each of the three scenarios was repeated three times in order to reduce the impact of random variations in robot navigation and perception, and to ensure the robustness of the results across repeated trials. Frames were extracted from the video streams at a constant rate of one frame per second (recommended for video processing [82]) resulting in the dataset summarized in Table 2. In total, Scenario A produced 147 frames across three runs, Scenario B produced 134 frames, and Scenario C produced 161 frames, leading to an overall dataset of 442 frames. This dataset served as the input for all tested models, with identical preprocessing and configuration to maintain consistency.

Table 2 Dataset details for Scenarios A, B, and C.

| Scenario | First Run | Second Run | Third Run | Total Frames |
|---|---|---|---|---|
| A | 49 | 47 | 51 | 147 |
| B | 46 | 45 | 43 | 134 |
| C | 51 | 54 | 56 | 161 |
| Total | | | | 442 |

Considering the nature of safety inspection reports, each frame can be evaluated in terms of safe or unsafe classification. This evaluation was based on standard classification metrics, true positives (TP), true negatives (TN), false positives (FP), and false negatives (FN), which allowed us to compute accuracy, precision, recall, and F1-scores. In the context of this research, true positives (TP) refer to frames where the framework correctly identified a safety violation that was present, while true negatives (TN) are frames correctly classified as safe when no violation existed. Conversely, false positives (FP) occur when the system incorrectly flagged a frame as unsafe despite the absence of any violation, and false negatives (FN) represent frames where actual violations were missed and labeled as safe. These metrics provide a evaluation of both detection accuracy and reasoning reliability, which is essential for safety-related applications in the AEC industry [83], [84]. The equations for these measures are given below:

(1) $$\text{Accuracy} = \frac{TP + TN}{TP + TN + FP + FN}$$

(2) $$\text{Precision} = \frac{TP}{TP + FP}$$

(3) $$\text{Recall} = \frac{TP}{TP + FN}$$

(4) $$\text{F1 score} = \frac{2 \times (\text{Precision} \times \text{Recall})}{\text{Precision} + \text{Recall}}$$

The objective is to show how well the proposed framework identified unsafe conditions in Scenarios A, B, and C and to provide a benchmark that others can reproduce. Because the final report layer involves summarization that compresses and reformats evidence, quantitative comparisons are made at the safety assessment module (i.e., the output after the second VLM pass described in Section 3.5), where each frame is labeled safe or unsafe and contributes to the TP/TN/FP/FN counts introduced above. We benchmark our system against GPT-4o for three reasons. (1) In our pre-evaluation on the study scenarios, GPT-4o outperformed the strongest open models we tested, Gemma-3n and MiniCPM-2.6V, by 14% and 21%, respectively. (2) GPT-4o is a native multimodal model (image/video/audio/text) designed for real-time perception-language reasoning, which aligns with our framework architecture; independent analyses also report GPT-4o as a top non-reasoning multimodal foundation model on standard vision tasks and relatively robust under distribution shift. (3) Recent peer-reviewed studies within the context of robotics and VLMs adopts GPT-4o as a comparator, enabling direct comparability to their proposed models, for example: PhysVLM reports [85] results against GPT-4o on embodied reasoning; PathEval [86] evaluates compare various VLMs against GPT-4o on robot plan assessment; and a robotics-focused benchmark [87] profiles GPT-4o alongside OpenVLA and Open-X-Embodiment manipulation tasks. Figure 13 shows confusion matrices for our framework across Scenario A (left), B (middle), and C (right). In Scenario A, the system correctly identified 71 of 78 unsafe frames, with relatively few unsafe frames being mislabeled as safe (n=6). Another interesting observation is that the relatively high number of FP frames, meaning that 16 frames from 70 frames were mistakenly labeled unsafe while they were safe. Scenario B shows a modest decrease, from %92 to %86, in classifying true unsafe scenes in the stream. This decline is coupled with almost same amount of errors in false alarm for calling safe ones as unsafe scene. The observed trend in Scenario B is intensified in the Scenario C, where the number of true unsafe frames dropped into %79, reflecting the added complexity and overlapping hazards introduced in that setup. This decline is coupled with the increase in the number of frames identified as unsafe while they were safe. The analysis of these patterns was detailed in the discussion section and here we note only highlight the visible differences. Moreover, Figure 14 reports the corresponding confusion matrices for GPT-4o under the same conditions and frame set. The baseline follows a similar A to C trend but shows higher error rates overall, particularly in Scenarios A and B, where our framework achieves more correct unsafe detections while generating fewer false unsafe labels. Performance narrows in Scenario C, where both approaches face increased difficulty. Based on this evaluation, we can conclude that our framework provides stronger and more stable detection behavior in the first two scenarios while remaining competitive in the most challenging one.

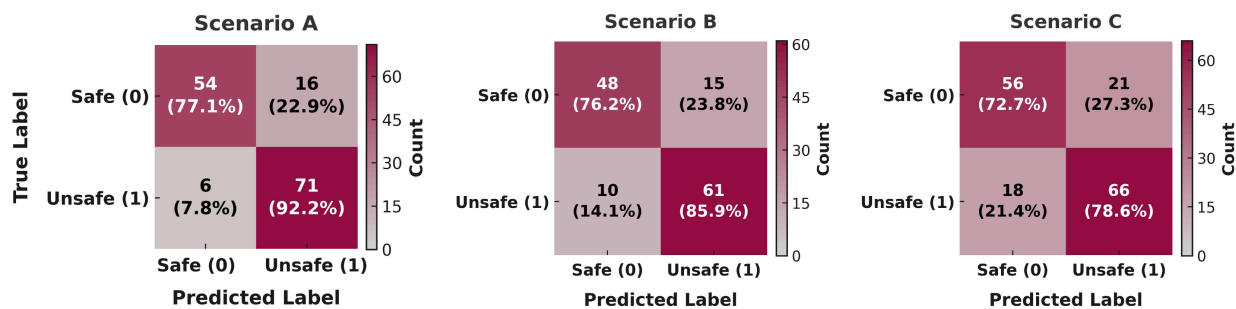

*Figure 13* Confusion matrix of results for the proposed framework

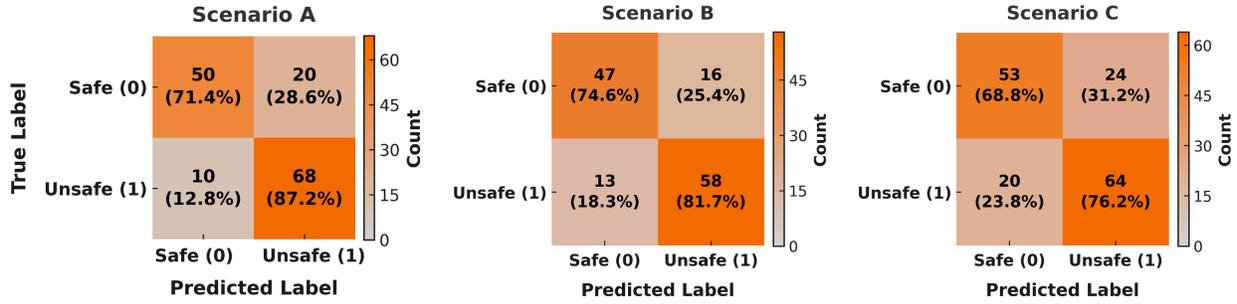

*Figure 14* Confusion matrix of results for the GPT-4o

Figure 15 summarizes the performance for our framework and GPT-4o across Scenarios A-C using the four standard metrics (recall, precision, accuracy, F1), allowing to evaluate the performance of the framework from different aspects and compare it with GPT4o. Generally, our framework showed more effectiveness in all metrics compared to GPT-4o. Another general observation is the downward trend from A to C, which can also be seen in GPT-4o's results and reflects the increasing complexity of scenarios with containing the overlap of hazards. It should be noted that while the performance difference between the proposed framework and GPT-4o is not big, the proposed framework is using open-source small foundation models that are almost 10 times cheaper than GPT-4o.

In particular, recall (see top-left part of diagram) measures how many of truly unsafe frames were correctly identified and reported through this inspection system. Our framework achieved 92.2% in A, 85.9% in B, and 78.6% in C, compared with GPT-4o's 87.2%, 81.7%, and 76.2%. The consistent margin between the framework and the GPT-4o model can be related to the effectiveness of adding generated rule into second pass of VLMs. Moreover, precision reflects how often frames flagged unsafe were truly unsafe. Both systems lose precision as scenarios get harder, largely due to more false identification of safe scenes as unsafe. The higher recall from precision means that the framework is more configured for the situations that identifying unsafe frames are preferred over false alarms. Accuracy captures overall correctness in all situations, whether it is safe or unsafe, while F1 summarized the balance between missed detections and false alarms. This means the framework detected most true violations while keeping false alarms low in ladder/housekeeping/electrical scenes. However, the lower F1 in Scenario C indicated reduced robustness in complex, overlapping-hazard and elevated-work settings led to more violation misses and more safe frames flagged as unsafe. It is important to note the principal role of recall metric in the case of safety inspection, and how the proposed system represents consistently high recall. In safety inspection, false negatives are the costliest mistakes because an unflagged hazard can lead directly to injury or death. So, it is more preferable to review an extra alert than to miss a real unsafe situation.

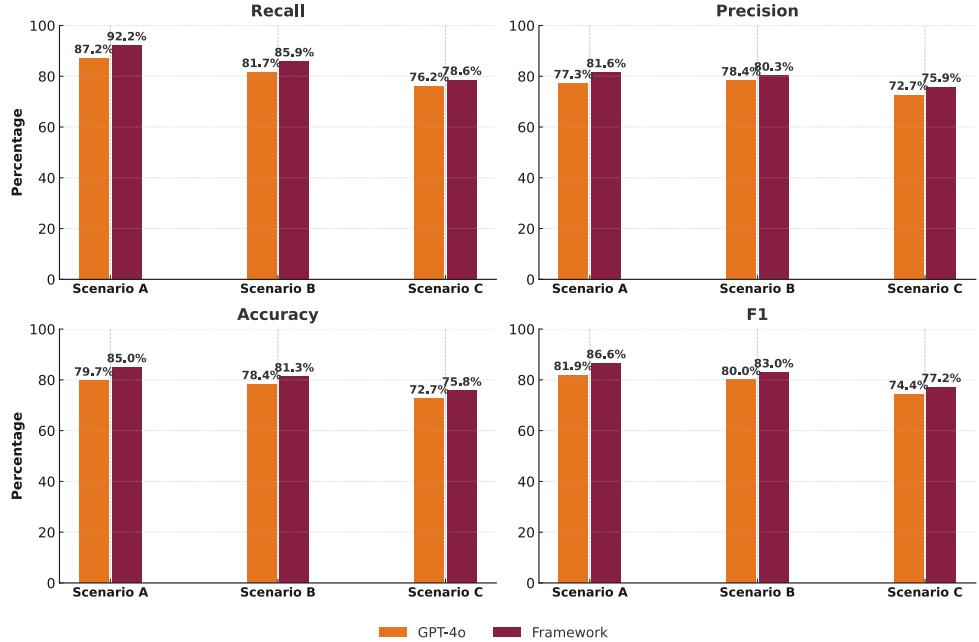

*Figure 15* Comparison of performance (recall, precision, accuracy, and F1 score) for the proposed framework with GPT4o

## 7. Discussion

This paper proposed a multi-layer framework, leveraging the state-of-the-art in robotic and foundation models to take one step closer to autonomy of safety management within the context of AEC industry. In section 6, we demonstrated the feasibility and performance of the proposed framework by implementing and comparing its results with other models. This section aims to contribute to the knowledge by providing an in-depth understanding of foundation model based autonomous system for construction safety management.

### 7.1. In-Depth Analysis of Output

This subsection goes deeper on one representative output so we can discuss how the system works end-to-end, and which parts can be further refined in future work. We used a labeling system in Figure 16 to connect the evidence together toward reaching a deeper insight later. Each label starts with B, which denotes the AI module. The first number is the frame index; the number after the dash is the component inside the module (1 = Perceptual Abstraction, 2 = Regulation Generation, 3 = Safety Assessment, 4 = Report Generation). The afterward letters show which upstream outputs were used to produce this output. For example, B16-3ABC shows frame 16, safety assessment (3), computed using prior A, B, and C outputs that are called out in the figure and then referenced forward.

The analysis proceeds in reverse pipeline order, beginning with the Safety Inspection Report, rooting its results back to the scene description. Entry 4.4 ("ladder misuse and PPE violations") aggregates several timestamps (e.g., 11 s, 12 s, 14-16 s, 42 s), and two of those are selected to go deeper (Frame 15 and Frame 16). In the Safety Assessment layer, B15-3A emphasizes the absence of a safety harness on the climber, while B16-3ABC focuses on PPE non-use and ladder position and stability. The difference in emphasis follows directly from the rules that the Regulation Generation component produced for each frame. In Frame 15, the rule set (see B15-2B) highlighted correct ladder use (three-point contact, facing the ladder, clear landing area) and flagged the harness as a concern under site-policy and height conditions. In Frame 16, the rule set (see B16-2BC) emphasized PPE requirements around the forklift areas and ladder setup (side-rail extension or securement and the 4:1 angle), so the assessment concentrated on the missing hard hat or vest and the not-fully-secured ladder. Moreover, those rules themselves were grounded in concrete scene evidence from the first layer. In Frame 16, B16-1B and B16-1C captured that a yellow hard hat and a high-visibility vest are on the floor, not worn by the worker on the ladder; the description also notes the extension ladder's state, nearby carts, and the proximity of the forklift. In Frame 15, B15-1A recorded the "Safety Gear" lying on the floor, the forklift dominating the right side, a fall-hazard sign on the mast, and a worker on the ladder. When the Regulation Generation layer reads these descriptions, it retrieves the relevant rules and composes them into the frame-specific checklist. The Safety Assessment then applies those rules to the same evidence and produces the frame-level judgment that ultimately feeds the report. Therefore, this traceable and transparent system not only allows each layer to generate more effective output but also helps to get involve users and human in the loop, as a move away from current "black box" AI systems [88].

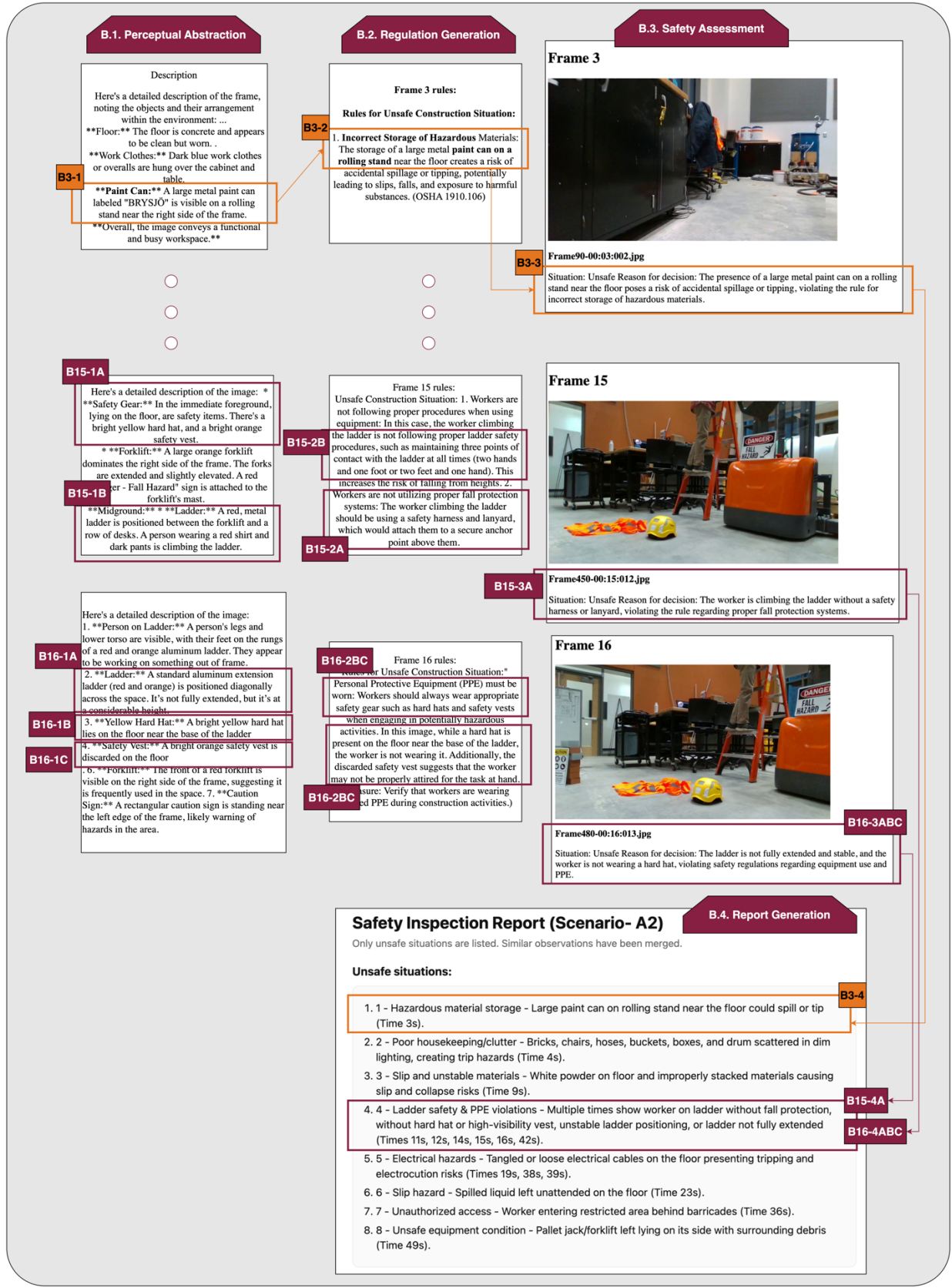

Figure 16 Detailed view to the sample output

This process also helps to better understand the limitation of the system and further improve it in next tests. In the figure, B3-4 contributes a hazardous-material storage entry to the report, but this is a false positive result (see more in confusion matrix), meaning that the framework labeled a safe situation as unsafe and generated a false alarm. Tracing back shows the cause: the first layer misidentified a large blue water container as a paint can, and the camera perspective made it look "on" a rolling stand when the stand was actually behind it. The Regulation Generation layer then correctly pulled a rule about improper storage of liquids on rolling stands, and the Safety Assessment (B3-3) carried that forward. The error, however, began with the description. This is a typical hallucination/grounding issue caused by look-alike objects and depth ambiguity. It points to two places that still need work: more robust object typing and spatial reasoning in the Perceptual Abstraction layer, and additional spatial sanity checks (e.g., depth/occlusion tests) before a storage rule is allowed to fire (see detailed in the section Limitations and Future Studies).

### 7.2. Significance

The paper's contributions toward enhanced robotic autonomy in construction jobsites are as follows. The first point of significance is transparency. Despite their invaluable computation and cognitive power, foundation models are often criticized as black boxes, which complicates adoption in regulated, safety-critical contexts [88]. Our layered pipeline addresses this by producing explainable and understandable artifacts at each layer. This modularity provides traceability, allowing us to understand which object or rule changed the outcome. This is aligned with best-practice calls from the foundation-model literature and explainable-AI research [89]. RAG unit further strengthens this transparency by tying language outputs to safety regulations rather than relying purely on training memory, reducing hallucinations and enabling auditable reasoning over up-to-date rulesets. In short, instead of a single rigid output (or report), the system takes verifiable steps toward a conclusion, enabling human-in-the-loop as a part of system.

Moreover, the framework is flexible and generalizable in two complementary ways. Architecturally, the multi-layer design lets researchers and practitioners change detectors, add task-specific heads, or plug in new retrieval sources without retraining the entire system. That is aligned with current multi-task and modular approaches in construction safety monitoring, which report gains by tailoring components while keeping the overall interface stable. Apart from the architectural level, our zero-shot, pre-trained framework facilitates the adoption of the framework in the AEC settings, where the lack of specific datasets is frequently mentioned as a common challenge for adoption of traditional computer vision models [90]. Reviews of PPE-compliance systems and open visual datasets in AEC repeatedly note data scarcity, domain shift, and limited generalization as barriers [91]. Additionally, the system can be deployed in privacy important manners, such as construction jobsites. Because components are lightweight and open-source compatible, they can run on edge devices or behind a site firewall, keeping images and inferences locally. This is aligned with recent research demonstrating that when firms do adopt AI safety monitoring, anonymization and clear data-handling rules are central to acceptance, reinforcing the contribution of our system's architecture [92], [93].

### 7.3. Limitations and Future Studies

Although the section 5 demonstrates the performance of the proposed framework, several critical limitations need to be addressed before this platform can be scaled and widely implemented. In this section, we explored the key limitations of the current system based on our observations and propose directions for future research that align with recent academic findings. First, although our SLAM-based robotic module supports fleet-style operation, it still depends on users defining navigation goals and waypoints. More autonomy still is required for continuous, round-the-clock exploration on dynamic construction sites. More research is needed to design autonomous exploration algorithms for the situations such as safety inspection that the robot should priorities unknown and most dangerous locations over simple paths. For example, Liu et al. [94]. present a frontier-based strategy that enables mobile robots to plan efficient exploration paths without manual guidance, improving coverage and mapping accuracy in unknown environments Moreover, in our own experiments, we observed that clutter, typical in construction zones, and vibrations at higher traversal speeds broke alignment and degraded map fidelity. Thus, future iterations should integrate autonomous frontier-planning SLAM with uncertainty-aware path selection to enhance resilience under environmental perturbation.

Beyond robotics, the AI module requires more work in following areas. While performance in detecting unsafe situations is promising, false alarms, particularly in unfamiliar or complex scenarios, remain an open research area. One core challenge impacting on this area is the hallucination behavior of foundation models, where they generate inaccurate reasoning or labels that are not validated and true (see B3-1 in Figure 17). Some research is going on to detect and mitigate hallucinations, which can undermine trust and safety of deploying foundation model-based frameworks [95]. Next, although our system improves on scene description and rule grounding, spatial understanding remains a limitation. Current VLMs encode text and vision in a shared space, but they often lack physical reasoning, a serious problem when distinguishing spatial relationships integral to safety (e.g., whether a ladder is properly secured, if a head is above a guardrail, or if an object obstructs escape routes). Studies reveal that VLMs still struggle with spatial deformation and relational reasoning, even on tasks as basic as "object A is left of B", with considerable gaps between human and model accuracy [96], [97].

Third, most VLMs process discrete frames rather than coherent video. However, safety inspection often depends on observing temporal context, sequences of movement, causality, and continuity, which static frames cannot capture. Although video-grounded models are emerging [55], [56], many still lack true temporally integrated reasoning, instead treating frame sequences as independent inputs. Recent efforts like Scene-R1 demonstrate video-grounded 3D scene reasoning without dense point-wise supervision by combining temporal snippet selection with 2D grounding and 3D projection [98]. A future version of our system can incorporate Scene-R1 architecture that can track objects over time and infer temporal safety patterns, such as a ladder swaying, a worker losing balance, or machinery shifting.

### 8. Conclusion

This paper aims to enhance the autonomy and productivity in safety inspections by moving it one step away from manual, or task-specific automated system toward a generalizable process that can be ran on a mobile robot. We introduced a multi-layer framework that turns raw site vision data streams into a traceable safety report: a VLM produces scene descriptions, a retrieval component grounds those descriptions in OSHA-based rules, an explicit safety assessment applies the rules, and a report layer explains what was found and why. On the robotic side, the system is built to increase autonomy so that the platform can navigate the environment, gather evidence systematically, and feed higher-quality inputs to the AI layers. We implemented a working prototype on a quadruped, integrated localization and mapping to enable waypoint-to-waypoint navigation and successfully evaluated the end-to-end pipeline across scenarios that reflect common hazards on jobsites.

The results show that the approach is viable in terms of safety violation detection and report generation. The system reached high recall across scenarios while keeping precision at a level that is manageable in workflows where missing a real hazard is costlier than reviewing an extra alert. The layered design also is a move toward more transparent and traceable AI. Because each layer produces an artifact, captions, rules, frame-level judgments, and final report, supervisors can audit the path from pixels to a sentence in the report. In our sample walk-through, the labels (e.g., B16-3ABC) made it easy to trace how a single observation flowed through the pipeline and where to fix an error when it appeared. The significance of this design is twofold. First, it aligns the metric profile with safety priorities: high recall to reduce missed hazards, supported by downstream reasoning to filter noise. Second, it replaces a single rigid decision with a chain of small, checkable steps, which can potentially increase the trust. It also creates natural points to integrate site policy, to add new checks, or to swap components without retraining the whole system. In contrast to narrow, task-specific computer vision modules, our zero-shot, retrieval approach adapts to new tasks and sites with much less data and setup time, while the edge-friendly architecture supports deployment where privacy and connectivity are real constraints, such as AEC jobsites.

The deep dive on one representative output illustrated how this traceability helps teams improve the system. We showed how final report entries come from frame-level assessments, which themselves come from rules tied to rules and supported by concrete visual evidence. We also showed how a false alarm could be explained and repaired: a first layer mislabel of a water container as paint led, step by step, to an incorrect hazardous-materials entry. Because each step is visible, we can target fixes rather than guessing at an end-to-end model's state. Thresholds can be tuned by hazard class; repeat findings across frames can be fused to avoid false alert and flagged items can be linked to corrective actions or work orders. On the organizational side, keeping data on site and providing citations eases concerns from workers and managers, while the clear artifact trail supports audits and incident reviews. Over time, the same system can power continuous improvement, turning false positives and near-misses into targeted updates to the retrieval set or the spatial checks.

At the same time, our study is not exception to limitations that not only gives a better understanding of foundation model applications in AEC but also shape next steps. Autonomy is still limited from robotic side, where the robot can localize and map, but it relies on user-provided goals, and map alignment can drift in clutter or under vibration situation. Moreover, spatial and temporal understanding remain open problems in this context, where many hazards need advanced spatial and temporal understandings, not just what they are in a single frame. Our future work targets these gaps with uncertainty-aware exploration, video-level reasoning and 3D reasoning, with lightweight components, and stronger traceability checks that guard against false alarms. Therefore, this paper contributes a multi-layer transparent framework for automated safety inspection, a working prototype on a mobile platform, and an evaluation that shows the metric performances. The system turns unstructured imagery into rule based, reviewable reports that safety teams can act on. We view this as a foundation to build a deeper autonomy for 24/7 coverage, richer spatial-temporal reasoning for complex scenes, and broader field trials across sites and seasons. With these advances, automated inspection can shift from promising demos to dependable practice in the AEC industry.

**Reference**

[1]     BLS, "Number and rate of fatal work injuries, by industry sector," U.S. BUREAU OF LABOR STATISTICS. Accessed: May 10, 2022. [Online]. Available: https://www.bls.gov/charts/census-of-fatal-occupational-injuries/number-and-rate-of-fatal-work-injuries-by-industry.htm
[2]     M. C. Auld, J. C. H. Emery, D. V. Gordon, and D. McClintock, "The Efficacy of Construction Site Safety Inspections," *J Labor Econ*, vol. 19, no. 4, pp. 900–921, Oct. 2001, doi: 10.1086/322826.
[3]     A. Albert, M. R. Hallowell, and B. M. Kleiner, "Enhancing Construction Hazard Recognition and Communication with Energy-Based Cognitive Mnemonics and Safety Meeting Maturity Model: Multiple Baseline Study," *J Constr Eng Manag*, vol. 140, no. 2, p. 04013042, Sep. 2013, doi: 10.1061/(ASCE)CO.1943-7862.0000790.
[4]     OSHA, "Competent Person ," Occupational Safety and Health Administration. Accessed: Sep. 24, 2024. [Online]. Available: https://www.osha.gov/competent-person
[5]     S. Bahn, "Workplace hazard identification and management: The case of an underground mining operation," *Saf Sci*, vol. 57, pp. 129–137, Aug. 2013, doi: 10.1016/J.SSCI.2013.01.010.
[6]     A. Carbonari, A. Giretti, and B. Naticchia, "A proactive system for real-time safety management in construction sites," *Autom Constr*, vol. 20, no. 6, pp. 686–698, Oct. 2011, doi: 10.1016/J.AUTCON.2011.04.019.
[7]     J. Park, E. Marks, Y. K. Cho, and W. Suryanto, "Performance Test of Wireless Technologies for Personnel and Equipment Proximity Sensing in Work Zones," *J Constr Eng Manag*, vol. 142, no. 1, p. 04015049, Jul. 2015, doi: 10.1061/(ASCE)CO.1943-7862.0001031.
[8]     S. Halder and K. Afsari, "Robots in Inspection and Monitoring of Buildings and Infrastructure: A Systematic Review," *Applied Sciences 2023, Vol. 13, Page 2304*, vol. 13, no. 4, p. 2304, Feb. 2023, doi: 10.3390/APP13042304.

[9] M. Gheisari and B. Esmaeili, "Applications and requirements of unmanned aerial systems (UASs) for construction safety," *Saf Sci*, vol. 118, pp. 230–240, Oct. 2019, doi: 10.1016/J.SSCI.2019.05.015.

[10] S. Halder, K. Afsari, E. Chiou, R. Patrick, and K. A. Hamed, "Construction inspection & monitoring with quadruped robots in future human-robot teaming: A preliminary study," *Journal of Building Engineering*, vol. 65, p. 105814, Apr. 2023, doi: 10.1016/J.JOBE.2022.105814.

[11] M. Gheisari, A. Rashidi, and B. Esmaeili, "Using Unmanned Aerial Systems for Automated Fall Hazard Monitoring," *Construction Research Congress 2018: Safety and Disaster Management - Selected Papers from the Construction Research Congress 2018*, vol. 2018-April, pp. 62–72, 2018, doi: 10.1061/9780784481288.007.

[12] I. H. Kim, H. Jeon, S. C. Baek, W. H. Hong, and H. J. Jung, "Application of Crack Identification Techniques for an Aging Concrete Bridge Inspection Using an Unmanned Aerial Vehicle," *Sensors 2018, Vol. 18, Page 1881*, vol. 18, no. 6, p. 1881, Jun. 2018, doi: 10.3390/S18061881.

[13] H. H. Wang and F. Boukamp, "Leveraging Project Models for Automated Identification of Construction Safety Requirements," *Congress on Computing in Civil Engineering, Proceedings*, pp. 240–247, 2007, doi: 10.1061/40937(261)30.

[14] J. Trogh, D. Plets, L. Martens, and W. Joseph, "Improved tracking by mitigating the influence of the human body," *2015 IEEE Globecom Workshops, GC Wkshps 2015 - Proceedings*, 2015, doi: 10.1109/GLOCOMW.2015.7414013.

[15] S. Mastrolembo Ventura, P. Bellagente, S. Rinaldi, A. Flammini, and A. L. C. Ciribini, "Enhancing Safety on Construction Sites: A UWB-Based Proximity Warning System Ensuring GDPR Compliance to Prevent Collision Hazards," *Sensors (Basel)*, vol. 23, no. 24, p. 9770, Dec. 2023, doi: 10.3390/S23249770.

[16] Y. K. Cho, J. H. Youn, and D. Martinez, "Error modeling for an untethered ultra-wideband system for construction indoor asset tracking," *Autom Constr*, vol. 19, no. 1, pp. 43–54, Jan. 2010, doi: 10.1016/J.AUTCON.2009.08.001.

[17] Y. Gong, J. O. Seo, T. W. Kim, S. Ahn, and Y. Luo, "Field validation of beacon-based indoor tracking and localization system for construction workers," *KSCE Journal of Civil Engineering*, vol. 29, no. 2, p. 100017, Feb. 2025, doi: 10.1016/J.KSCEJ.2024.100017.

[18] B. W. Jo, Y. S. Lee, J. H. Kim, D. K. Kim, and P. H. Choi, "Proximity Warning and Excavator Control System for Prevention of Collision Accidents," *Sustainability 2017, Vol. 9, Page 1488*, vol. 9, no. 8, p. 1488, Aug. 2017, doi: 10.3390/SU9081488.

[19] S. Halder and K. Afsari, "Robots in Inspection and Monitoring of Buildings and Infrastructure: A Systematic Review," *Applied Sciences 2023, Vol. 13, Page 2304*, vol. 13, no. 4, p. 2304, Feb. 2023, doi: 10.3390/APP13042304.

[20] I. Jeelani and M. Gheisari, "Safety challenges of UAV integration in construction: Conceptual analysis and future research roadmap," *Saf Sci*, vol. 144, p. 105473, Dec. 2021, doi: 10.1016/J.SSCI.2021.105473.

[21] A. Ghadimzadeh Alamdari, F. A. Zade, and A. Ebrahimkhanlou, "A Review of Simultaneous Localization and Mapping for the Robotic-Based Nondestructive Evaluation of Infrastructures," *Sensors 2025, Vol. 25, Page 712*, vol. 25, no. 3, p. 712, Jan. 2025, doi: 10.3390/S25030712.

[22] M. F. Ahmed, K. Masood, V. Fremont, and I. Fantoni, "Active SLAM: A Review on Last Decade," *Sensors 2023, Vol. 23, Page 8097*, vol. 23, no. 19, p. 8097, Sep. 2023, doi: 10.3390/S23198097.

[23] G. Younes, D. Asmar, E. Shammas, and J. Zelek, "Keyframe-based monocular SLAM: design, survey, and future directions," *Rob Auton Syst*, vol. 98, pp. 67–88, Dec. 2017, doi: 10.1016/J.ROBOT.2017.09.010.

[24] Y. R. Lee, S. H. Jung, K. S. Kang, H. C. Ryu, and H. G. Ryu, "Deep learning-based framework for monitoring wearing personal protective equipment on construction sites," *J Comput Des Eng*, vol. 10, no. 2, pp. 905–917, Mar. 2023, doi: 10.1093/JCDE/QWAD019.

[25] M. I. B. Ahmed *et al.*, "Personal Protective Equipment Detection: A Deep-Learning-Based Sustainable Approach," *Sustainability 2023, Vol. 15, Page 13990*, vol. 15, no. 18, p. 13990, Sep. 2023, doi: 10.3390/SU151813990.

[26] Y. S. Shin and J. Kim, "A Vision-Based Collision Monitoring System for Proximity of Construction Workers to Trucks Enhanced by Posture-Dependent Perception and Truck Bodies' Occupied Space," *Sustainability 2022, Vol. 14, Page 7934*, vol. 14, no. 13, p. 7934, Jun. 2022, doi: 10.3390/SU14137934.

[27] D. Kim, M. Liu, S. H. Lee, and V. R. Kamat, "Remote proximity monitoring between mobile construction resources using camera-mounted UAVs," *Autom Constr*, vol. 99, pp. 168–182, Mar. 2019, doi: 10.1016/J.AUTCON.2018.12.014.

[28] M. Gheisari, B. Esmaeili, J. Kosecka, and A. Rashidi, "Using Unmanned Aerial Systems for Automated Fall Hazard Monitoring in High-rise Construction Projects," 2020.

[29] Y. G. Lim, J. Wu, Y. M. Goh, J. Tian, and V. Gan, "Automated classification of 'cluttered' construction housekeeping images through supervised and self-supervised feature representation learning," *Autom Constr*, vol. 156, p. 105095, Dec. 2023, doi: 10.1016/J.AUTCON.2023.105095.

[30] W. Fang *et al.*, "Computer vision applications in construction safety assurance," *Autom Constr*, vol. 110, p. 103013, Feb. 2020, doi: 10.1016/J.AUTCON.2019.103013.

[31] S. V. T. Tran *et al.*, "Leveraging large language models for enhanced construction safety regulation extraction," *ITcon Vol. 29, Special issue Managing the digital transformation of construction industry (CONVR 2023), pg. 1026-1038, http://www.itcon.org/2024/45*, vol. 29, no. 45, pp. 1026–1038, Dec. 2024, doi: 10.36680/J.ITCON.2024.045.

[32] H. Naderi and A. Shojaei, "Large-Language model (LLM)-Powered system for Situated and Game-Based construction safety training," *Expert Syst Appl*, vol. 283, p. 127887, Jul. 2025, doi: 10.1016/J.ESWA.2025.127887.

[33] F. Yang and J. Zhang, "Prompt-based automation of building code information transformation for compliance checking," *Autom Constr*, vol. 168, p. 105817, Dec. 2024, doi: 10.1016/J.AUTCON.2024.105817.

[34] M. Adil, G. Lee, V. A. Gonzalez, and Q. Mei, "Using Vision Language Models for Safety Hazard Identification in Construction," Apr. 2025, Accessed: Aug. 19, 2025. [Online]. Available: https://arxiv.org/pdf/2504.09083

[35] Y. Wang, B. Xiao, A. Bouferguene, and M. Al-Hussein, "Proactive safety hazard identification using visual–text semantic similarity for construction safety management," *Autom Constr*, vol. 166, p. 105602, Oct. 2024, doi: 10.1016/J.AUTCON.2024.105602.

[36] A. Diwan, L. Berry, E. Choi, D. Harwath, and K. Mahowald, "Why is Winoground Hard? Investigating Failures in Visuolinguistic Compositionality," *Proceedings of the 2022*

*Conference on Empirical Methods in Natural Language Processing, EMNLP 2022*, pp. 2236–2250, 2022, doi: 10.18653/V1/2022.EMNLP-MAIN.143.

[37] K. Morman, A. Costin, and J. Mcnair, "ENABLING INTELLIGENT CONSTRUCTION: CURRENT CHALLENGES AND CONSIDERATIONS FOR THE CONNECTED SITE".

[38] C. Rudin, "Stop explaining black box machine learning models for high stakes decisions and use interpretable models instead," *Nat Mach Intell*, vol. 1, no. 5, pp. 206–215, May 2019, doi: 10.1038/S42256-019-0048-X;SUBJMETA=117,4002,4014,4045,531,639,705;KWRD=COMPUTER+SCIENCE,CRIMINOLOGY,SCIENCE.

[39] A. Birk and M. Pfingsthorn, "Simultaneous Localization and Mapping (SLAM)," *Wiley Encyclopedia of Electrical and Electronics Engineering*, pp. 1–24, Nov. 2016, doi: 10.1002/047134608X.W8322.

[40] C. Cadena *et al.*, "Past, Present, and Future of Simultaneous Localization And Mapping: Towards the Robust-Perception Age," *IEEE Transactions on Robotics*, vol. 32, no. 6, pp. 1309–1332, Jan. 2017, doi: 10.1109/TRO.2016.2624754.

[41] S. Thrun and M. Montemerlo, "The Graph SLAM Algorithm with Applications to Large-Scale Mapping of Urban Structures," *Int J Rob Res*, vol. 25, no. 5–6, pp. 403–429, May 2006, doi: 10.1177/0278364906065387.

[42] K. Karur, N. Sharma, C. Dharmatti, and J. E. Siegel, "A Survey of Path Planning Algorithms for Mobile Robots," *Vehicles 2021, Vol. 3, Pages 448-468*, vol. 3, no. 3, pp. 448–468, Aug. 2021, doi: 10.3390/VEHICLES3030027.

[43] M. Iovino, E. Scukins, J. Styrud, P. Ögren, and C. Smith, "A survey of Behavior Trees in robotics and AI," *Rob Auton Syst*, vol. 154, p. 104096, Aug. 2022, doi: 10.1016/J.ROBOT.2022.104096.

[44] P. Lewis *et al.*, "Retrieval-Augmented Generation for Knowledge-Intensive NLP Tasks," *Adv Neural Inf Process Syst*, vol. 2020-December, May 2020, Accessed: Feb. 13, 2024. [Online]. Available: https://arxiv.org/abs/2005.11401v4

[45] W. Fan *et al.*, "A Survey on RAG Meeting LLMs: Towards Retrieval-Augmented Large Language Models," *Proceedings of the ACM SIGKDD International Conference on Knowledge Discovery and Data Mining*, pp. 6491–6501, May 2024, doi: 10.1145/3637528.3671470.

[46] W. Zhou, S. Zhang, H. Poon, and M. Chen, "Context-faithful Prompting for Large Language Models," *Findings of the Association for Computational Linguistics: EMNLP 2023*, pp. 14544–14556, Mar. 2023, doi: 10.18653/v1/2023.findings-emnlp.968.

[47] H. Cui *et al.*, "CURIE: Evaluating LLMs on Multitask Scientific Long-Context Understanding and Reasoning".

[48] A. Elnashar, J. White, and D. C. Schmidt, "Enhancing structured data generation with GPT-4o evaluating prompt efficiency across prompt styles," *Front Artif Intell*, vol. 8, p. 1558938, Mar. 2025, doi: 10.3389/FRAI.2025.1558938/BIBTEX.

[49] M. Labbé and F. Michaud, "RTAB-Map as an Open-Source Lidar and Visual SLAM Library for Large-Scale and Long-Term Online Operation," *J Field Robot*, vol. 36, no. 2, pp. 416–446, Mar. 2024, doi: 10.1002/rob.21831.

[50] J. Zhang, Y. Yao, and B. Deng, "Fast and Robust Iterative Closest Point," *IEEE Trans Pattern Anal Mach Intell*, vol. 44, no. 7, pp. 3450–3466, Apr. 2023, doi: 10.1109/TPAMI.2021.3054619.

[51] S. Macenski, F. Martín, R. White, and J. G. Clavero, "The Marathon 2: A Navigation System," *IEEE International Conference on Intelligent Robots and Systems*, pp. 2718–2725, Jul. 2020, doi: 10.1109/IROS45743.2020.9341207.

[52] S. Macenski, T. Moore, D. V. Lu, A. Merzlyakov, and M. Ferguson, "From the desks of ROS maintainers: A survey of modern & capable mobile robotics algorithms in the robot operating system 2," *Rob Auton Syst*, vol. 168, Oct. 2023, doi: 10.1016/j.robot.2023.104493.

[53] ollama, "Ollama." Accessed: Aug. 12, 2025. [Online]. Available: https://ollama.com/

[54] H. Liu, C. Li, Q. Wu, and Y. J. Lee, "Visual Instruction Tuning," *Adv Neural Inf Process Syst*, vol. 36, Apr. 2023, Accessed: Sep. 23, 2024. [Online]. Available: https://arxiv.org/abs/2304.08485v2

[55] Y. Yao *et al.*, "MiniCPM-V: A GPT-4V Level MLLM on Your Phone," Aug. 2024, Accessed: Sep. 23, 2024. [Online]. Available: https://arxiv.org/abs/2408.01800v1

[56] G. Team *et al.*, "Gemma 3 Technical Report," Mar. 2025, Accessed: Aug. 12, 2025. [Online]. Available: https://arxiv.org/pdf/2503.19786

[57] A. Q. Jiang *et al.*, "Mistral 7B," Oct. 2023, Accessed: Sep. 23, 2024. [Online]. Available: https://arxiv.org/abs/2310.06825v1

[58] H. Touvron *et al.*, "Llama 2: Open Foundation and Fine-Tuned Chat Models," Jul. 2023, Accessed: Sep. 23, 2024. [Online]. Available: https://arxiv.org/abs/2307.09288v2

[59] G. Team *et al.*, "Gemma 2: Improving Open Language Models at a Practical Size," Jul. 2024, Accessed: Sep. 23, 2024. [Online]. Available: https://arxiv.org/abs/2408.00118v2

[60] M. Abdin *et al.*, "Phi-3 Technical Report: A Highly Capable Language Model Locally on Your Phone," Apr. 2024, Accessed: Sep. 23, 2024. [Online]. Available: https://arxiv.org/abs/2404.14219v4

[61] A. Dubey *et al.*, "The Llama 3 Herd of Models," Jul. 2024, Accessed: Sep. 23, 2024. [Online]. Available: https://arxiv.org/abs/2407.21783v2

[62] DeepSeek-AI *et al.*, "DeepSeek-R1: Incentivizing Reasoning Capability in LLMs via Reinforcement Learning," Jan. 2025, Accessed: Aug. 12, 2025. [Online]. Available: https://arxiv.org/pdf/2501.12948

[63] P. Sahoo, A. K. Singh, S. Saha, V. Jain, S. Mondal, and A. Chadha, "A Systematic Survey of Prompt Engineering in Large Language Models: Techniques and Applications," Feb. 2024, Accessed: Aug. 12, 2025. [Online]. Available: https://arxiv.org/pdf/2402.07927

[64] S. Vatsal and H. Dubey, "A Survey of Prompt Engineering Methods in Large Language Models for Different NLP Tasks," Jul. 2024, Accessed: Aug. 12, 2025. [Online]. Available: https://arxiv.org/pdf/2407.12994

[65] R. Duan, H. Deng, M. Tian, Y. Deng, and J. Lin, "SODA: A large-scale open site object detection dataset for deep learning in construction," *Autom Constr*, vol. 142, Oct. 2022, doi: 10.1016/j.autcon.2022.104499.

[66] J. Lee and S. Lee, "Construction Site Safety Management: A Computer Vision and Deep Learning Approach," *Sensors 2023, Vol. 23, Page 944*, vol. 23, no. 2, p. 944, Jan. 2023, doi: 10.3390/S23020944.

[67] Y. Sun, I. Jeelani, and M. Gheisari, "Safe human-robot collaboration in construction: A conceptual perspective," *J Safety Res*, vol. 86, pp. 39–51, Sep. 2023, doi: 10.1016/J.JSR.2023.06.006.

[68] T. P. Huck, C. Ledermann, and T. Kroger, "Testing Robot System Safety by creating Hazardous Human Worker Behavior in Simulation," *IEEE Robot Autom Lett*, vol. 7, no. 2, pp. 770–777, Nov. 2021, doi: 10.1109/LRA.2021.3133612.

[69] J. Lee and S. Lee, "Construction Site Safety Management: A Computer Vision and Deep Learning Approach," *Sensors*, vol. 23, no. 2, Jan. 2023, doi: 10.3390/S23020944.

[70] A. Stocco, B. Pulfer, and P. Tonella, "Mind the Gap! A Study on the Transferability of Virtual vs Physical-world Testing of Autonomous Driving Systems," *IEEE Transactions on Software Engineering*, vol. 49, no. 4, pp. 1928–1940, Aug. 2022, doi: 10.1109/TSE.2022.3202311.

[71] C. Camargo, J. Gonçalves, M. Conde, F. J. Rodríguez-Sedano, P. Costa, and F. J. García-Peñalvo, "Systematic Literature Review of Realistic Simulators Applied in Educational Robotics Context," *Sensors (Basel)*, vol. 21, no. 12, p. 4031, Jun. 2021, doi: 10.3390/S21124031.

[72] OSHA, "1926.1053 - Ladders. ," Occupational Safety and Health Administration. Accessed: Aug. 14, 2025. [Online]. Available: https://www.osha.gov/laws-regs/regulations/standardnumber/1926/1926.1053

[73] OSHA, "1926.25 - Housekeeping," Occupational Safety and Health Administration. Accessed: Aug. 14, 2025. [Online]. Available: https://www.osha.gov/laws-regs/regulations/standardnumber/1926/1926.25

[74] OSHA, "Electrical - Construction," Occupational Safety and Health Administration. Accessed: Aug. 14, 2025. [Online]. Available: https://www.osha.gov/electrical/construction

[75] OSHA, "Personal Protective Equipment - Construction ," Occupational Safety and Health Administration. Accessed: Aug. 14, 2025. [Online]. Available: https://www.osha.gov/personal-protective-equipment/construction

[76] OSHA, "Construction Focus Four," *OSHA Directorate of Training and Education*, 2011.

[77] S. Harwood, "Safety Training for Focus Four Hazards in Construction," *OSHA*, 2006.

[78] C. Socias-Morales, "The Problem of Falls from Elevation in Construction and Prevention Resources," NIOSH. Accessed: Aug. 14, 2025. [Online]. Available: https://blogs.cdc.gov/niosh-science-blog/2024/05/01/falls-2024/

[79] BLS, "Fatal injuries from ladders down in 2020; nonfatal ladder injuries were essentially unchanged ," U.S. Bureau of Labor Statistics. Accessed: Aug. 14, 2025. [Online]. Available: https://www.bls.gov/opub/ted/2022/fatal-injuries-from-ladders-down-in-2020-nonfatal-ladder-injuries-were-essentially-unchanged.htm

[80] K. Sapungan, "Top 7 Construction Accidents: Causes, Prevention, and Real Examples," mastt. Accessed: Aug. 14, 2025. [Online]. Available: https://www.mastt.com/blogs/construction-accidents

[81] M. Shridhar, L. Manuelli, and D. Fox, "CLIPort: What and Where Pathways for Robotic Manipulation," *Proc Mach Learn Res*, vol. 164, pp. 894–906, Sep. 2021, Accessed: Jun. 08, 2024. [Online]. Available: https://arxiv.org/abs/2109.12098v1

[82] Google, "Video analysis with Gemma ," Google AI for Developers. Accessed: Aug. 15, 2025. [Online]. Available: https://ai.google.dev/gemma/docs/capabilities/vision/video-understanding

[83] L. Messi, A. Corneli, M. Vaccarini, and A. Carbonari, "Development of a Twin Model for Real-time Detection of Fall Hazards," *Proceedings of the 37th International Symposium on Automation and Robotics in Construction, ISARC 2020: From Demonstration to

*Practical Use - To New Stage of Construction Robot*, pp. 256–263, 2020, doi: 10.22260/ISARC2020/0037.

[84] N. Rane, S. Choudhary, and J. Rane, "Integrating Leading-edge Sensors for Enhanced Monitoring and Controlling in Architecture, Engineering and Construction: A Review," *SSRN Electronic Journal*, Nov. 2023, doi: 10.2139/SSRN.4644138.

[85] W. Zhou *et al.*, "PhysVLM: Enabling Visual Language Models to Understand Robotic Physical Reachability," Mar. 2025, Accessed: Sep. 03, 2025. [Online]. Available: https://arxiv.org/pdf/2503.08481

[86] M. Y. Aghzal, "Evaluating Vision-Language Models as Evaluators in Path Planning," CVPR. Accessed: Sep. 03, 2025. [Online]. Available: https://cvpr.thecvf.com/virtual/2025/poster/32727?utm_source=chatgpt.com

[87] P. Guruprasad *et al.*, "Benchmarking Vision, Language, & Action Models on Robotic Learning Tasks," Dec. 2024, Accessed: Sep. 03, 2025. [Online]. Available: https://arxiv.org/pdf/2411.05821v1

[88] R. Bommasani *et al.*, "On the Opportunities and Risks of Foundation Models," Aug. 2021, Accessed: Mar. 27, 2024. [Online]. Available: https://arxiv.org/abs/2108.07258v3

[89] S. Ali *et al.*, "Explainable Artificial Intelligence (XAI): What we know and what is left to attain Trustworthy Artificial Intelligence," *Information Fusion*, vol. 99, p. 101805, Nov. 2023, doi: 10.1016/J.INFFUS.2023.101805.

[90] A. M. Vukicevic, M. Petrovic, P. Milosevic, A. Peulic, K. Jovanovic, and A. Novakovic, "A systematic review of computer vision-based personal protective equipment compliance in industry practice: advancements, challenges and future directions," *Artif Intell Rev*, vol. 57, no. 12, pp. 1–28, Dec. 2024, doi: 10.1007/S10462-024-10978-X/FIGURES/7.

[91] R. Xiong *et al.*, "OpenConstruction: A Systematic Synthesis of Open Visual Datasets for Data-Centric Artificial Intelligence in Construction Monitoring," Aug. 2025, Accessed: Aug. 18, 2025. [Online]. Available: https://arxiv.org/pdf/2508.11482

[92] M. Gugssa, L. Li, L. Pu, A. Gurbuz, Y. Luo, and J. Wang, "Enhancing the Time Efficiency of Personal Protective Equipment (PPE) Detection in Real Implementations Using Edge Computing," *Computing in Civil Engineering 2023: Resilience, Safety, and Sustainability - Selected Papers from the ASCE International Conference on Computing in Civil Engineering 2023*, pp. 532–540, 2024, doi: 10.1061/9780784485248.064;PAGE:STRING:ARTICLE/CHAPTER.

[93] C. Reaño, J. V. Riera, V. Romero, P. Morillo, and S. Casas-Yrurzum, "A cloud-edge computing architecture for monitoring protective equipment," *Journal of Cloud Computing 2024 13:1*, vol. 13, no. 1, pp. 1–17, Apr. 2024, doi: 10.1186/S13677-024-00649-1.

[94] C. Liu *et al.*, "Enhancing autonomous exploration for robotics via real time map optimization and improved frontier costs," *Sci Rep*, vol. 15, no. 1, pp. 1–12, Dec. 2025, doi: 10.1038/S41598-025-97231-9;SUBJMETA=166,639,987;KWRD=ELECTRICAL+AND+ELECTRONIC+ENGINEERING,ENGINEERING.

[95] L. Huang *et al.*, "A Survey on Hallucination in Large Language Models: Principles, Taxonomy, Challenges, and Open Questions," *ACM Trans Inf Syst*, vol. 43, no. 2, Jan. 2025, doi: 10.1145/3703155.

[96] J. Zhang *et al.*, "Ascending the Infinite Ladder: Benchmarking Spatial Deformation Reasoning in Vision-Language Models," Jul. 2025, Accessed: Aug. 19, 2025. [Online]. Available: https://arxiv.org/pdf/2507.02978
[97] F. Liu, G. Emerson, and N. Collie, "Visual Spatial Reasoning," *Trans Assoc Comput Linguist*, vol. 11, pp. 635–651, Apr. 2022, doi: 10.1162/tacl_a_00566.
[98] Z. Yuan *et al.*, "Scene-R1: Video-Grounded Large Language Models for 3D Scene Reasoning without 3D Annotations," Jun. 2025, Accessed: Aug. 19, 2025. [Online]. Available: https://arxiv.org/pdf/2506.17545v1